\documentclass[journal]{IEEEtran}
\usepackage{tabularx}
\usepackage{soul}
\usepackage{url}
\usepackage[small]{caption}
\usepackage{graphicx}
\usepackage{float}
\usepackage{amsmath}
\usepackage{booktabs}
\usepackage{subfigure}
\usepackage{algorithm}
\usepackage{color}
\usepackage{algorithmic}
\usepackage{amssymb}
\usepackage{amsmath}
\usepackage{multirow}
\usepackage{comment}
\usepackage{makecell}
\usepackage{bm}
\usepackage{hyperref}
\usepackage{mathrsfs}

\begin{document}
%
\title{Subspace-Contrastive Multi-View Clustering}

\author{Lele~Fu, Lei~Zhang, Jinghua Yang, Chuan Chen*, Chuanfu Zhang, and Zibin Zheng, \emph{Senior Member}, IEEE

\thanks{Lele Fu and Chuanfu Zhang are with the School of System Sciences and Engineering, Sun Yat-sen University, Guangzhou, China. Lei Zhang, Chuan Chen, and Zibin Zheng are with the School of Computer Science and Engineering, Sun Yat-sen University, Guangzhou, China. Jinghua Yang is with Faculty of Information Technology, Macau University of Science and Technology, Macau, China. (email: lawrencefzu@gmail.com, chenchuan@mail.sysu.edu.cn). * Corresponding author.
}}

\maketitle

\begin{abstract}
Most multi-view clustering methods are limited by shallow models without sound nonlinear information perception capability, or fail to effectively exploit complementary information hidden in different views. To tackle these issues, we propose a novel Subspace-Contrastive Multi-View Clustering (SCMC) approach. Specifically, SCMC utilizes view-specific auto-encoders to map the original multi-view data into compact features perceiving its nonlinear structures.
Considering the large semantic gap of data from different modalities, we employ subspace learning to unify the multi-view data into a joint semantic space, namely the embedded compact features are passed through multiple self-expression layers to learn the subspace representations, respectively.
In order to enhance the discriminability and efficiently excavate the complementarity of various subspace representations, we use the contrastive strategy to maximize the similarity between positive pairs while differentiate negative pairs. Thus, a weighted fusion scheme is developed to initially  learn a consistent affinity matrix. Furthermore,  we employ the graph regularization to encode the local geometric structure within varying subspaces for further fine-tuning the appropriate affinities between instances. To demonstrate the effectiveness of the proposed model, we conduct a large number of comparative experiments on eight challenge datasets, the experimental results show that SCMC outperforms existing shallow and deep multi-view clustering methods.
\end{abstract}


\IEEEpeerreviewmaketitle

\section{Introduction}
With the growing popularity of data generation and feature extraction, multi-view or multimedia data are available in large quantities.
To be specific, multi-view data refer to various feature representations from multiple aspects of objects. For instance, an image can be characterized by wavelet texture (WT), local binary pattern (LBP), histogram of oriented gradient (HOG), etc. A piece of document can be expressed in numerous languages.
Researchers generally believe that multi-view data consist of rich and useful heterogeneous information,
so the technologies related to multi-view analysis  are receiving increasing attention.
Multi-view clustering (MVC) \cite{FurecentMulti2020,zhao2017multilearning,zong2020Multimodal} is one of the representative technologies, which aims to explore the complementary and consistent information embedded in multi-view data to boost the clustering performance.

Currently, there are extensive multi-view clustering methods. For example, graph-based MVC \cite{wang2020GMC,wen2021Adaptive,wen2021Structural} learned the connectivity graph matrices to reveal the relationship of samples, then the designed fusion schemes are developed to merge these graph matrices into a global graph. Spectral embedding-based MVC \cite{Li2021Consensus,hajjar2022constrained,Hajjar2022spectral} exploited low-dimensional spectral embedding with orthogonal constraint for each view, which portrays important components of data, a consensus representation was further learned on the basis of them.
The goal of nonnegative matrix based MVC \cite{xu2020nonnegativeembedding,yang2021Orthogonal,huang2022Factorization} is to factorize a nonnegative discrete cluster indicator matrix from varying representations, thus the $argmax(\cdot)$ function is adopted to acquire the data labels.
Among multitudinous MVC methods, multi-view subspace clustering is a research hotspot and widely studied for its superior performance, which absorbs theory from conventional subspace clustering and further develops it.
The works \cite{gao2015Subspace,zhang2020generalized} are classic multi-view subspace clustering approaches, which aimed to explore a uniform underlying feature space from multiple subspace representations.
\cite{li2022Correlation,chen2021low} performed the tensor factorization on the representation tensor to capture the global correlations between views.
These shallow models have yielded promising clustering results, but most real-world data are high-dimensional and nonlinear, shallow models might not been equipped with the ability to fetch nonlinear structures.

Auto-Encoder (AE) is an effective unsupervised deep representation learning paradigm, which non-linearly maps the original data features into a compact feature space via the encoders, then passes the compact representations through the decoders to reconstruct the data.
AE is frequently used to condense data information in clustering tasks.
\cite{xie2016Embedding,guo2017improved} are two well-known deep embedding learning methods, which used Kullback-Leibler (KL) divergence regularization to maximize the similarity of soft assignments and target distributions.
During the past few years, AE is also introduced to multi-view subspace clustering. Sun et al. \cite{sun2019self} used self-supervised strategy to improve the unified subspace representation learning.
Zhu et al. \cite{zhu2019networks} simultaneously learned a set of view-specific self-expression representations, then which are combined into a common self-expression representation.
Wang et al. \cite{wang2021Deep} learned a unified subspace representation from multi-view discriminative  feature spaces.
Cui et al. \cite{cui2021Guided} proposed the spectral supervisor to guide the learning of consensus subspace representation.
The clustering performance of the above deep multi-view subspace clustering approaches are excellent, but their abilities of exploiting the association between multiple subspace representations still need to be improved.
For instance, \cite{sun2019self,wang2021Deep} directly learn the consistent self-expression representation from multi-view latent features refined by AEs, which could not capture the characteristics of disparate views, thus failing to utilizing the complementary information.
\cite{zhu2019networks} applies a Hilbert Schmidt Independence Criterion (HSIC) regularization term to reinforce the diversity of different views, this indistinguishable alienation of different views may render it difficult to obtain the agreement of them.
As for \cite{cui2021Guided}, a weighted fusion layer is used to integrate all self-expression representations, which does not harness the view correlations in insightful ways.
Contrastive learning \cite{grill2020bootstrap} is an emerging self-supervised strategy that aims to maximize the similarity between positive pairs whereas minimize the similarity between negative pairs.
In multi-view clustering scenarios, there is a natural contrastive relationship between varying views, thus giving rise to some multi-view contrastive clustering methods \cite{hassani2020contrastive,xu2022Multilevel,trosten2021reconsidering}.
An important objective reality in multi-view data is that there may be large modality gap of data under different views, which can drive the distance between instance pairs to be extremely huge, rendering the contrast process difficult. The current multi-view contrastive clustering methods barely consider and address
the problem of modality gap, this is a vital motivation of this paper.

\begin{figure}[htbp]
	\centering
	\includegraphics[width=0.4\textwidth]{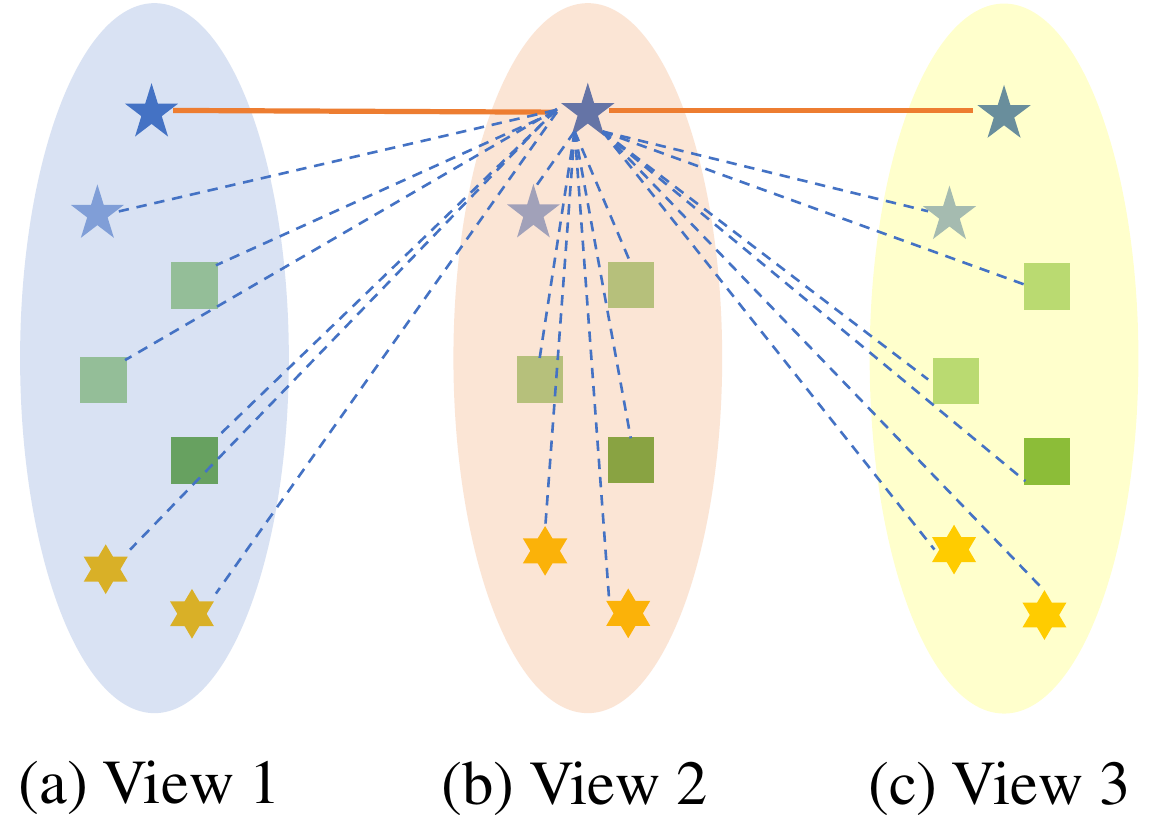}
	\caption{The diagram of constructing the positive and negative pairs. Let us take three views as an example, the first data point of view 2 and the same data points of the other two views are positive pairs (connected by solid lines). The first data point of view 2 and the remaining data points of the other two views and its own view are negative pairs (connected by dashed lines).}
	\label{contrast-diagram}
\end{figure}

We are inspired by the idea of contrastive learning, and propose a Subspace-Contrastive
Multi-View Clustering (SCMC) method.
Specifically, in order to perceive the nonlinear structures in multi-view data, we employ view-specific AEs to encode the initial features into multiple compact space, wherein the respective subspace representations are further learned through the self-expression layers, such that the semantic information of data belonging to disparate modalities can be unified into a common semantic space.
Thus, we consider the same sample under different views as the positive pairs, and the rest of pairs are considered as negative, Fig. \ref{contrast-diagram} illustrates the manner of constructing positive and negative pairs.
By pairwisely contrasting multiple subspace representations, we bring the positive pairs closer together and the negative pairs further apart.
This operation enhances the discriminability of each subspace representation and explore the complementary information within them, which is different from the discrimination-induced regularization achieved by the indistinguishable mutual exclusion between various representations in literatures \cite{wang2021Deep,zhu2019networks}.
To obtain a consistent affinity matrix, we use a weighted fusion scheme to merge multiple subspace representations.
Moreover, the graph regularization is applied to encode the local structures inside the learned subspaces.
Finally, abundant experiments on eight challenge datasets are implemented to verify the effectiveness of the proposed SCMC. The major contributions of this paper are summarized as follows:
\begin{itemize}
  \item We adopt view-specific auto-encoders to map the multi-view data into compact feature spaces perceiving the nonlinear structures, wherein respective subspace representations are explored via self-expression layers to align the semantic information of data under diverse modalities into a unified semantic space.
  \item We consider different subspace representations as contrast targets, then the pairwise contrast of them is conducted to exploit the complementarity between heterogeneous views and enhance the discrimination of each subspace representation.
  \item To demonstrate the validity of the proposed SCMC, we carry out comprehensive experiments on eight multi-view datasets, and the experimental results show that SCMC possesses advanced data clustering capability compare with the baseline and other multi-view clustering methods.
\end{itemize}

The rest of this paper are structured as follows. Section \ref{related_work} briefly review the works related to multi-view subspace clustering and contrastive learning. In Section \ref{SCMC}, we explicate the objective loss and the network architecture of SCMC. Experimental details are narrated in Section \ref{Experiments}. Finally, conclusion is summarized in Section \ref{conclusion}.

\begin{figure*}[htbp]
	\centering
	\includegraphics[width=1\textwidth]{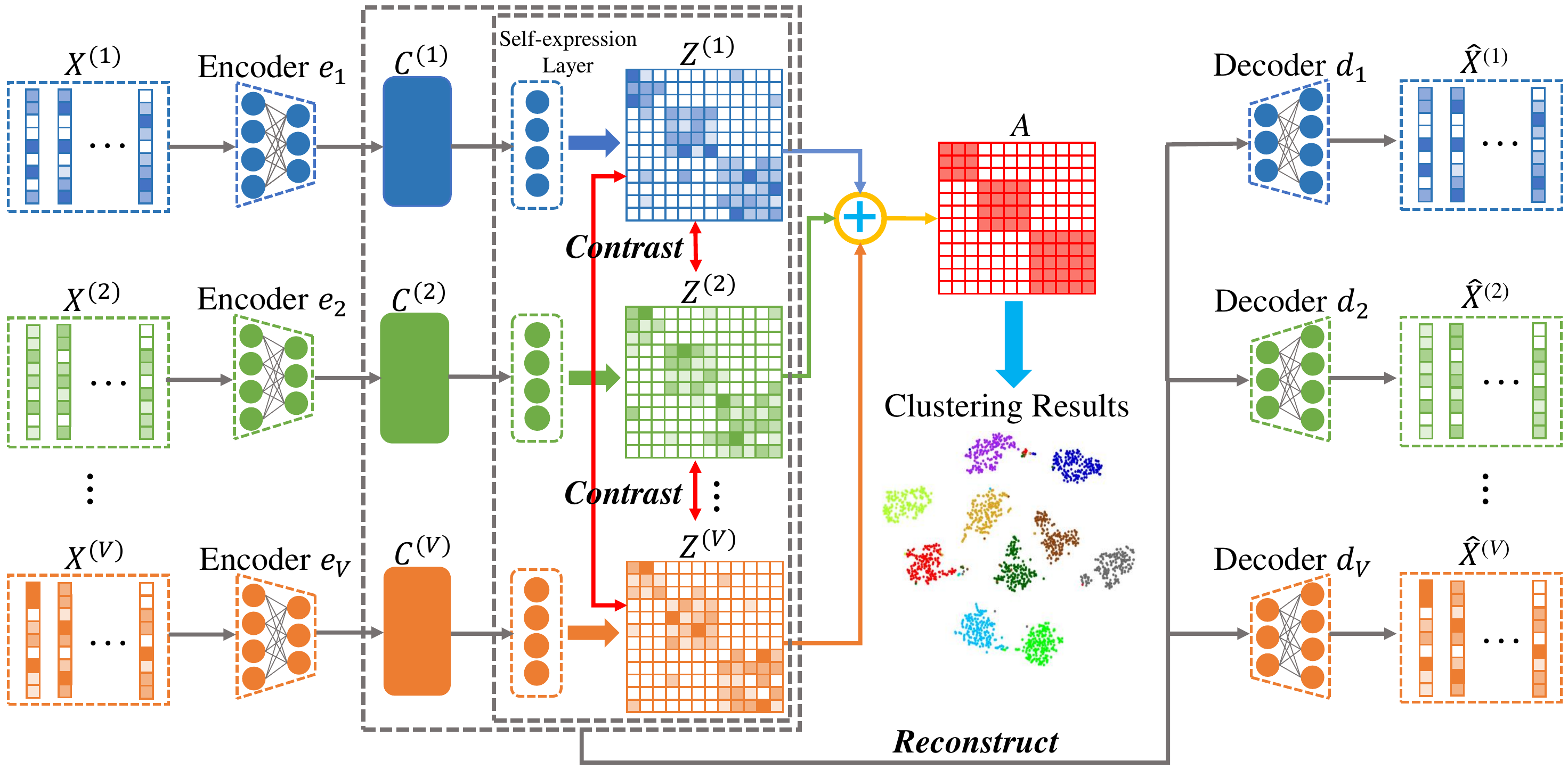}
	\caption{The framework of the proposed SCMC. To effectively handle high-dimensional and nonlinear structures in data, we use $V$ view-specific encoders to encode the initial multi-view features $\{\mathbf{X}^{(v)}\}_{v=1}^{V}$ as the compact embedding features $\{\mathbf{C}^{(v)}\}_{v=1}^{V}$. Thus, $\{\mathbf{C}^{(v)^{T}}\}_{v=1}^{V}$ pass through the multiple self-expression layers to obtain the features $\{\mathbf{C}^{(v)^{T}}\mathbf{Z}^{(v)}\}_{v=1}^{V}$, which are fed into the $V$ view-specific decoders to reconstruct the recovered data $\{\hat{\mathbf{X}}^{(v)}\}_{v=1}^{V}$. Notably, $\{\mathbf{Z}^{(v)}\}_{v=1}^{V}$ are essentially the coefficient matrices of multiple self-expression layers, also called the subspace representations. We contrast these subspace representations in pairs to exploit the complementary information between them. Additionally, a weighted fusion of all subspace representations is performed to obtain a unified affinity matrix while the graph regularization is adopted to fine-tune the affinities. Finally, the spectral algorithm is employed to acquire the clustering results. }
	\label{SCMC-framework}
\end{figure*}

\section{Related Works}\label{related_work}
\subsection{Multi-view Subspace Clustering}
Multi-view subspace clustering leverages heterogeneous features of data to group samples into a union of diverse subspaces.
Self-expression based subspace learning technology have gained widespread attention due to its concise but sound feature characterization capabilities.
Extensive Self-expression based multi-view subspace clustering approaches have also been proposed. For instance, \cite{zhang2020generalized,luo2018consistent,liu2021Incomplete} aimed to explore a common subspace representation from multi-source features, which is further imposed on certain designed regularization terms to fit clustering task. For capturing the high-order correlations among views, \cite{chen2021low,xie2020hyper,li2022preserved} reorganized multiple self-expression matrices into a third-order tensor, which is constrained with the tensor nuclear norm to recover a low-rank feature space.
To decrease the high computational complexity caused by subspace learning, \cite{Kang2020Large,wang2022Guidance} introduced the anchor graph embedding to approximate the global affinity matrices. As the above methods are based on shallow models, their abilities to match the complex data distributions are limited.
For bridge this gap, \cite{zhu2019networks,wang2021Deep} used deep network frameworks to learn the subspace representations, then they all adopted an exclusive regularization term to improve the independence of each view, thus further boosting the complementary information among varying views. However, the exclusive regularization terms may also magnify the variability between positive pairs, making it difficult to obtain a uniform representation with high confidence.
\subsection{Contrastive Learning}
Contrastive learning is one of the research hot spots of the self-supervised learning paradigm over recent years, its core idea is to bring the distance between positive pairs closer and the distance between negative pairs farther in a projection feature space.
In practice, \cite{he2020momentum,chen2020simple,niu2021Semantic} were proposed successively in computer vision filed, which enhanced the discrimination of image representations by means of contrastive learning to better serve the downstream tasks.
Owing to the presented favorable performance, contrastive learning have gained attention and been applied in clustering field.  Li et al. \cite{li2021Contrastive} simultaneously contrasted the
instance-level and cluster-level representations to strength the separability of samples belonging to different clusters.
Zhong et al. \cite{zhong2021Graph} learned clustering-friendly features and  compact clustering assignments via the designed contrastive strategy.
Furthermore, researchers have extended single-view contrastive clustering to multi-view cases,
Pan et al. \cite{pan2021Graph} proposed a multi-view contrastive graph nodes clustering scheme, wherein a consistent graph is learned via the graph contrastive loss. Xu et al. \cite{xu2022Multilevel} conducted data reconstruction in a low-level space and punished the consistent objectives via a contrastive scheme in a high-level space. In light of view invariance and local structure, Trosten et al. \cite{trosten2021reconsidering} proposed a selective contrastive alignment method to address the misaligned label distributions.
\section{The Proposed Method}\label{SCMC}
In this section, we first explain the  motivations for proposing the SCMC method. Second, we
present the objective functions of the primary modules in the proposed SCMC, revealing the mathematical details and the effect behind them. Thus, the specific network architectures are summarized, which are also graphically illustrated in Fig. \ref{SCMC-framework} for better comprehension. Finally, the training process is explained.
\subsection{Motivation}
\begin{enumerate}
  \item[1)] Traditional shallow multi-view subspace clustering methods \cite{gao2015Subspace,luo2018consistent,li2022Correlation} are limited in perceiving the nonlinear structures in multi-view data. Therefore, we want to utilize deep neural networks to handle this problem, thus allowing data with complex distributional properties to separate well in subspaces.
  \item[2)] Existing multi-view subspace clustering methods \cite{zhu2019networks,wang2021Deep} enhance the discrimination of different representations through a exclusive regularization term, but this undifferentiated disparity of all representations may render the models difficult to acquire the agreement across views. Inspired by contrastive learning, we differentially treat samples from various views, construct cross-view positive and negative pairs, and strengthen the discrimination of subspace representations by bringing positive pairs closer and separating negative pairs.
  \item[3)] The semantic information gap between different modalities in multi-view data \cite{xu2013survey} could be very large. For example, the HOG feature of an image describes completely different semantic information from the LBP feature, then the corresponding feature representations are extremely disparate. Most current multi-view clustering methods based on contrastive learning fail to effectively address this issue, making the contrast between sample pairs very difficult. In light of this, we explore the subspace representations of all views to unify the semantic information from heterogeneous views into a joint semantic space, facilitating to perform the contrast between sample pairs.
\end{enumerate}

\subsection{Objective Function}
\emph{1) Reconstruction and Subspace Losses}:
A multi-view dataset is denoted by $\{\mathbf{X}^{(v)}\}_{v=1}^{V}$, where $\mathbf{X}^{(v)}\in \mathbb{R}^{N\times d^{(v)}}$, $N$, $d^{(v)}$ represent the number of instances and the dimension of the $v$-th view's feature matrix. For the original feature $\mathbf{X}^{(v)}$, it may be high-dimensional and nonlinear, which poses difficulties for the downstream tasks. Hence, we use multiple view-specific encoders to nonlinearly map $\mathbf{X}^{(v)}$ into a latent low-dimensional space. For the $v$-th view, its encoder is formulated as
\begin{equation}\label{encode}
   \mathbf{C}^{(v)} = \mathcal{e}_{v}(\mathbf{X}^{(v)}|\mathbf{W}^{(v)}_{e}, \mathbf{b}^{(v)}_{e}),
\end{equation}
where $\mathbf{C}^{(v)}$ is the embedding feature after $\mathbf{X}^{(v)}$ passing the encoder $\mathcal{e}_{v}(\cdot)$, $\mathbf{W}^{(v)}_{e}$ and $\mathbf{b}^{(v)}_{e}$ indicate the weight matrix and bias vector in the encoder, respectively.

Current multi-view contrastive clustering  tends to first project multi-view data into compact feature spaces and then perform contrast between them.
However, the semantic information of data from heterogeneous views can be very disparate. Even if some works such as \cite{xu2022Multilevel} considers projecting data from different views into a unified feature space through a shared network, the distance between pairs could still be very large. It is difficult to pull together the positive pairs of different modalities.
Subspace representation is not only an informative low-dimensional representation form of data, but it also contains an important property, i.e., it portrays the affinity relationship between sample points.
For example, given a subspace representation $\mathbf{Z}\in \mathbb{R}^{N\times N}$, $\mathbf{Z}_{ij}$ measures the affinity between the $i$-th instance and the $j$-th instance. Thus, the $i$-th row $\mathbf{Z}_{i}$ can be considered as a low-dimensional subspace representation of the $i$-th instance, but also as the affinities of the $i$-th instance with other instances. Thus, there are still differences in the subspace representations from diverse views, but their semantic information remains consistent, alleviating the dilemma in the contrast process. We learn the subspace representation $\mathbf{Z}^{(v)}$ of the $v$-th view by
\begin{equation}\label{subspace}
   \min_{\mathbf{C}^{(v)},\mathbf{Z}^{(v)}}||\mathbf{C}^{(v)^{T}}-\mathbf{C}^{(v)^{T}}\mathbf{Z}^{(v)}||_{F}^{2}.
\end{equation}
In the networks, $\mathbf{Z}^{(v)}$ is coefficient matrix of the learnable self-expression layer, which is achieved via one-layer fully connected layer without the bias part.
After the embedding feature $\mathbf{C}^{(v)^{T}}$ passes through the self-expression layer $\mathbf{Z}^{(v)}$ to obtain $\mathbf{C}^{(v)^{T}}\mathbf{Z}^{(v)}$, we reconstruct the data via feeding it into the decoder $\mathcal{d}_{v}(\cdot)$, the decoding process is formulated as
\begin{equation}\label{decode}
   \hat{\mathbf{X}}^{(v)}=\mathcal{d}_{v}(\mathbf{C}^{(v)^{T}}\mathbf{Z}^{(v)}|\mathbf{W}_{d}^{(v)},\mathbf{b}_{d}^{(v)}),
\end{equation}
where $\hat{\mathbf{X}}^{(v)}$ denotes the reconstructed data, $\mathbf{W}_{d}^{(v)}$ and $\mathbf{b}_{d}^{(v)}$ represent the coefficient matrix and the bias vector of the decoder network, respectively. For $V$ views, the reconstruction loss and subspace representation learning loss are computed by Eq. \eqref{loss_reconstruction} and Eq. \eqref{loss_subspace}.

\begin{equation}\label{loss_reconstruction}
   \mathcal{L}_{Re}=\min_{\hat{\mathbf{X}}^{(v)}}\sum_{v=1}^{V}||\mathbf{X}^{(v)}-\hat{\mathbf{X}}^{(v)}||_{F}^{2}
\end{equation}

\begin{equation}\label{loss_subspace}
   \mathcal{L}_{Sub}=\min_{\mathbf{C}^{(v)},\mathbf{Z}^{(v)}}\sum_{v=1}^{V}||\mathbf{C}^{(v)^{T}}-\mathbf{C}^{(v)^{T}}\mathbf{Z}^{(v)}||_{F}^{2}.
\end{equation}

\emph{2) Contrastive Loss:}
There is a natural contrast between multiple subspace representations $\{\mathbf{Z}^{(v)}\}_{v=1}^{V}$. To exploit this property, we first construct positive and negative pairs across views. For $\mathbf{Z}^{(v)}_{i}$, it mutually forms a positive pair with the same instances under different views, while its negative samples contain all the instances except the positive samples. Fig. \ref{contrast-diagram} provides a graphical illustration of how positive and negative pairs are constructed.
Summarily, there are $V-1$ positive instances and $V(N-1)$ negative instances for $\mathbf{Z}^{(v)}_{i}$.
Thus, we apply consine distance to measure the similarity between pairs, the mathematical form is expressed as follows
\begin{equation}
\Theta\left(\mathbf{Z}_{i}^{(v_{1})}, \mathbf{Z}_{j}^{(v_{2})}\right)=\frac{\left(\mathbf{Z}_{i}^{(v_{1})}\right)\left(\mathbf{Z}_{j}^{(v_{2})}\right)^{\top}}{\left\|\mathbf{Z}_{i}^{(v_{1})}\right\|\left\|\mathbf{Z}_{j}^{(v_{2})}\right\|}
\end{equation}
Taking one view as an example, to achieve the goal of narrowing the positive pairs and widening the negative pairs, we formulate the problem as
\begin{equation}\label{loss_contrastive}
\begin{split}
     & \ell_{v}=-\min_{\mathbf{Z}^{(v)},\mathbf{Z}^{(k)}}\sum_{k=1,k\neq v}^{V}\sum_{i=1}^{N} \\
     & \!\!\!\!\!\log \frac{\exp \left(\Theta(\mathbf{Z}_{i}^{(v)}, \mathbf{Z}_{i}^{(k)}) / \tau\right)}{\sum_{j=1}^{N}\!\left(\!\exp (\Theta(\mathbf{Z}_{i}^{(v)}\!\!, \mathbf{Z}_{j}^{(v)}\!) / \tau)\!+\!\exp (\Theta(\mathbf{Z}_{i}^{(v)}\!\!, \mathbf{Z}_{j}^{(k)}\!) / \tau)\!\right)},
\end{split}
\end{equation}
where $\tau$ denotes the temperature parameter. We can observe that the numerator is about the calculation of positive pairs, while the denominator is about the calculation of negative pairs. The contrastable loss of $V$ views is computed by
\begin{equation}\label{loss_contrastive}
   \mathcal{L}_{Con}=\min\frac{1}{NV}\sum_{v=1}^{V}\ell_{v}
\end{equation}

\emph{3) Fusion Loss:}
For aggregating the complementary information in different views, we integrate multiple subspace representations into a consistent affinity matrix in a weighted fusion manner. Specifically, a set of weight coefficients are assigned to varying views, and they can be adaptively optimized in the back propagation process. The problem is written as
\begin{equation}\label{fusion}
\begin{split}
     & \mathbf{A} = \mathcal{f}(\{\mathbf{Z}^{(v)}\}_{v=1}^{V}|\Omega)=\sum_{v=1}^{V}\omega^{(v)}\mathbf{Z}^{(v)}\\
     & \text{s.t. } \sum_{v=1}^{V}\omega^{(v)}=1, \omega^{(v)} \geq 0,
\end{split}
\end{equation}
where $\mathcal{f}(\cdot)$ denotes the fusion function, $\Omega$ represents the learnable coefficients in the fusion function. In addition, an important assumption of graph embedding theory \cite{yan2006graph,nie2014projected} is that two samples closer to each other in the original space retain this property in the new low-dimensional space. We follow this assumption and consider that two instances similar in subspaces under any view, they should have higher affinity in the unified space. Thus, the following minimum problem can be obtained
\begin{equation}\label{loss_fusion}
  \begin{split}
       & \mathcal{L}_{Fu}=\min_{\mathbf{Z}^{(v)},\mathbf{A}}\sum_{v}\sum_{i}\sum_{j}||\mathbf{Z}_{i}^{(v)}-\mathbf{Z}_{j}^{(v)}||_{2}^{2}\mathbf{A}_{ij}
       +\sum_{i}\sum_{j}\mathbf{A}_{ij}^{2} \\
       & \quad  =\sum_{v=1}^{V}Tr(\mathbf{Z}^{(v)}\mathbf{L}_{\mathbf{A}}\mathbf{Z}^{(v)^{T}})+||\mathbf{A}||_{F}^{2}\\
       & \text{s.t. } \mathbf{A}_{i}\mathbf{1}=1, \mathbf{A}_{ij}\geq 0, \mathbf{A}_{ii}=0,
  \end{split}
\end{equation}
where $\mathbf{A}_{ij}$ is the ($i$, $j$)-th element in the uniform affinity matrix $\mathbf{A}$, $\mathbf{A}_{i}$ denotes the $i$-the row of $\mathbf{A}$, $\mathbf{1}$ is a vector with all elements of 1.
The constraints on $\mathbf{A}$ aim to avoid the trivial solutions. One may think that a unified affinity $\mathbf{A}$ can be learned directly through Eq. \eqref{loss_fusion}, and the weighted fusion mechanism seems to be unnecessary. Our intension is to initially obtain a consistent affinity matrix $\mathbf{A}$ through Eq. \eqref{fusion} to avoid that all elements of $\mathbf{A}$ are zeros, which makes Eq. \eqref{loss_fusion} unable to optimize, then we use Eq. \eqref{loss_fusion} to further fine-tune the affinities between samples.
At present, we give the final objective function of the proposed SCMC, which is written as
\begin{equation}\label{loss_final}
   \begin{split}
        & \mathcal{L}=\mathcal{L}_{Re}+\gamma_{1}\mathcal{L}_{Sub}+\gamma_{2}\mathcal{L}_{Con}+\gamma_{3}\mathcal{L}_{Fu} \\
        & =\min\sum_{v=1}^{V}\left(||\mathbf{X}^{(v)}-\hat{\mathbf{X}}^{(v)}||_{F}^{2}+\gamma_{1}||\mathbf{C}^{(v)}-\mathbf{C}^{(v)}\mathbf{Z}^{(v)}||_{F}^{2}\right)\\
        & -\gamma_{2}\sum_{v=1}^{V}\sum_{k=1,k\neq v}^{V}\sum_{i=1}^{N}\\
        &log \frac{\exp \left(\Theta(\mathbf{Z}_{i}^{(v)}, \mathbf{Z}_{i}^{(k)}) / \tau\right)}{\sum_{j=1}^{N}\!\left(\!\exp (\Theta(\mathbf{Z}_{i}^{(v)}\!\!, \mathbf{Z}_{j}^{(v)}\!) / \tau)\!+\!\exp (\Theta(\mathbf{Z}_{i}^{(v)}\!\!, \mathbf{Z}_{j}^{(k)}\!) / \tau)\!\right)}\\
        & +\gamma_{3}\sum_{v=1}^{V}Tr(\mathbf{Z}^{(v)}\mathbf{L}_{\mathbf{A}}\mathbf{Z}^{(v)^{T}}) +\gamma_{3}||\mathbf{A}||_{F}^{2},\\
        & \text{s.t. }\sum_{j=1}^{N}\mathbf{A}_{ij}=1, \mathbf{A}_{ij}\geq 0, \mathbf{A}_{ii}=0.
   \end{split}
\end{equation}
where $\gamma_{1}$, $\gamma_{2}$, and $\gamma_{3}$ are three nonnegative trade-off parameters. After the optimization based on back propagation, the consistent affinity $\mathbf{A}$ is obtained, we perform the spectral clustering algorithm on the matrix $(\mathbf{A}+\mathbf{A}^{T})/2$ to get the data labels.

\subsection{Network Architecture}
Based on the introduction of the objective function above, we sketch the network architecture of the proposed SCMC herein. $V$ three-layer encoders embed multi-view data $\{\mathbf{X}^{(v)}\}_{v=1}^{V}$ into compact features $\{\mathbf{C}^{(v)}\}_{v=1}^{V}$. Next, $\{\mathbf{C}^{(v)^{T}}\}_{v}^{V}$ pass the $V$ self-expression layers to obtain the matrix $\{\mathbf{C}^{(v)^{T}}\mathbf{Z}^{(v)}\}_{v=1}^{V}$, respectively.
Concretely, the self-expression layer is achieved by a one-layer linear layer discarding the bias part. Then,
$\{\mathbf{C}^{(v)^{T}}\mathbf{Z}^{(v)}\}_{v=1}^{V}$ is fed into $V$ decoders symmetrical to the encoders' structures to decode the reconstructed data $\{\hat{\mathbf{X}}^{(v)}\}_{v=1}^{V}$, respectively. In the encoding and decoding processes, the activation function $tanh(\cdot)$ is adopted. Furthermore, we contrast the learned subspace representations $\{\mathbf{Z}^{(v)}\}_{v=1}^{V}$ with each other, and fuse them into a consistent affinity matrix $\mathbf{A}$ with a group of learnable weights, then the nonnegative $\mathbf{A}$ is obtained via $Relu(\cdot)$ function. For fine-tuning the affinities between instances, graph regularization is leveraged to protect the local structures within multiple subspace representations.
\subsection{Training Details}
The training process of the network is divided in two steps: Pre-training and overall training. First, we pre-train $V$ three-layer AEs to initial their parameters. The purpose of pre-training is to mitigate the difficulties of training the overall network caused by all zeros in the AEs' parameters and the possibility of generating trivial solutions. Second, we train the overall network, the parameters of $V$ AEs, $V$ self-expression layers, a group of learnable weights, and the uniform affinity matrix are iteratively optimized.
Adam is adopted as the optimizer, and the learning rate is set to 0.0001. The experiments are run on a server equipped with Intel(R) Core(TM) i9-10980XE, RTX 3090 GPU, and 128G RAM.
Algorithm \ref{Algorithm_flow} provides a summary of the main steps of the proposed SCMC.
\begin{algorithm}
	\renewcommand{\algorithmicrequire}{\textbf{Input:}}
	\renewcommand{\algorithmicensure}{\textbf{Output:}}
	\caption{Subspace-Contrastive Multi-View Clustering}
	\label{Algorithm_flow}
	\begin{algorithmic}[1]
		\REQUIRE Multi-view data $\left\{\mathbf{X}^{(v)}\right\}_{v=1}^{V}$, parameters $\gamma_{1}$, $\gamma_{2}$, $\gamma_{3}$, and number of clusters $c$.	
		\ENSURE Consistent affinity matrix $\mathbf{A}$.
		\STATE Initialize multiple view-specific AEs with the parameters after pre-training, initialize the learning rate to 0.0001, the training epochs to 500.
		\FOR{$epoch$ = $1$ to $training$ $epochs$ }
		\STATE Compute the reconstruction loss $\mathcal{L}_{Re}$ by Eq. \eqref{loss_reconstruction};
        \STATE Compute the subspace learning loss $\mathcal{L}_{Sub}$ by Eq. \eqref{loss_subspace};
        \STATE Compute the contrast loss $\mathcal{L}_{Con}$ by Eq. \eqref{loss_contrastive};
        \STATE Obtain the initial consistent affinity matrix $\mathbf{A}$ by Eq. \eqref{fusion};
        \STATE Compute the local structures loss $\mathcal{L}_{Fu}$ by Eq. \eqref{loss_fusion};
        \STATE Compute the overall objective loss $\mathcal{L}$ by Eq. \eqref{loss_final} and update the network parameters via back propagation;
		\ENDFOR

        \STATE Performing the spectral clustering algorithm on $\frac{(\mathbf{A}+\mathbf{A}^{T})}{2}$ to acquire the data labels.
	\end{algorithmic}
\end{algorithm}
\section{Experiments} \label{Experiments}
\subsection{Multi-view Datasets}
Table \ref{description_of_data} summarizes the statistics of
eight test multi-view datasets. Specifically,
\textbf{ALOI}\footnote{https://elki-project.github.io/datasets/multi-view} contains 1,079 images from 10 objects, 4 features are extracted from these images: Haralick texture feature, HSV color histograms, Color
similarities, and RGB color histograms.
\textbf{GRAZ02}\footnote{http://www.emt.tugraz.at/˜pinz/data/GRAZ\_02} is object categorization dataset, which is composed of 1,476 images with 4 kinds of objects. 6 visual features are extracted, including Wavelet texture, Local binary pattern, SIFT feature, pyramid HOG, SURF feature, GIST feature. \textbf{NUS-WIDE}\footnote{https://lms.comp.nus.edu.sg/wp-content/uploads/2019/research/nuswide} is a famous image database, we select two subsets from it.
\textbf{NUS-WIDE-v1} contains 1,600 images of 8 categories, 6 views are Color histograms, Edge direction histograms, Block-wise color moments, Bag of words, Color correlograms, Wavelet textures, respectively. \textbf{NUS-WIDE-v2} consists of 2,000 images from 31 objects, each image are represented from 5 features: Color Histogram, Edge distribution, Color correlation,  Color moments, and Wavelet texture.
\textbf{Reuters}\footnote{ https://archive.ics.uci.edu/ml/datasets.html}
contains 1,500 documents with 5 languages, each sample is represented as a bag of words that extracted by the TFIDF-based weighting means.
\textbf{UCI} \cite{asuncion2007uci} is comprised of 2,000 handwritten numeric images ranging in [0, 9], three views are FOU feature, PIX feature, and MOR feature, respectively.
\textbf{WikipediaArticles}\footnote{http://lig-membres.imag.fr/grimal/data.html} is a document dataset organized by editors, which contains 693 short articles with 10 classes and 2 views.
\textbf{Youtube}\footnote{http://archive.ics.uci.edu/ml/datasets} is a multi-view video games dataset with 2,000 instances divided into 10 classes. Each entry has 6 features including HOG feature, CH feature, MFCC feature, VS feature, SS feature, and HME feature.
\begin{table*}[htbp]
	\centering
	\setlength{\tabcolsep}{2mm}{
		\caption{Statistics of eight datasets.}
		\vspace{0pt}
		\centering
		\begin{tabular}{|c|c|c|c|c|c|}
			\toprule
		     Dataset  & Views & Samples & Clusters & Features  \\
			\midrule
            ALOI & 4 & 1,079 & 10 & 64 / 64 / 77 / 13 \\
            GRAZ02 & 6 & 1,476 & 4 & 512 / 32 / 256 / 500 / 500 / 680  \\
            NUS-WIDE-v1 & 6 & 1,600 & 8 & 64 / 144 / 73 / 128 / 225 / 500  \\
            NUS-WIDE-v2  & 5 & 2,000 & 31 & 65 / 226 / 145 / 74 / 129  \\
            Reuters  & 5 & 1,500 & 6 & 21,531 / 24,892 / 34,251 / 15,506 / 11,547  \\
            UCI & 3 & 2,000 & 10 & 240 / 76 / 6  \\
            WikipediaArticles & 2 & 693 & 10 & 128 / 10  \\
            Youtube  & 6 & 2,000 & 10 & 2,000 / 1,024 / 64 / 512 / 64 / 647  \\
           \bottomrule
		\end{tabular}
		\label{description_of_data}}
\end{table*}

\subsection{Baseline and Compared Multi-view Methods}
To illustrate the validity of the proposed SCMC, we select the K-means algorithm as the baseline method. In practice, multiple features are concatenated together to form a unified feature representation, then it is fed into K-means to obtain the clustering labels. In addition, we collect ten state-of-the-art multi-view clustering approaches as the compared methods, which are briefly introduced as follows.
\begin{itemize}
  \item \textbf{CSMSC} \cite{luo2018consistent} learned the consensus subspace representation while the specific representations of different views are also explored.
  \item \textbf{MCGC} \cite{Zhan2019consensus} designed a disagreement cost function to reinforce the consensus among different graphs.
  \item \textbf{SM$^2$SC} \cite{Yang2019Split} pursued the view consistence via a variable splitting and a multiplicative decomposition module.
  \item \textbf{MvDSCN} \cite{zhu2019networks} proposed the diversity and uniformity networks to capture the view-specific and consistent information, respectively.
  \item \textbf{LMVSC} \cite{Kang2020Large} used anchor graph embedding to fit the global affinity matrix, thus resulting in linear computational complexity.
  \item \textbf{CGL} \cite{Li2021Consensus} optimized the spectral embedding matrices in a low-rank tensor space.
  \item \textbf{DMSC-UDL} \cite{wang2021Deep} learned a unified subspace representation while enhance the discrimination between diverse views.
  \item \textbf{EOMSC-CA} \cite{liu2022Efficient} fused the anchor graph scheme and graph construction into a uniform model.
  \item \textbf{CoMSC} \cite{liu2021Cotraining} employed the eigendecomposition to obtain the robust representations of multi-view data, from which the consistent self-expressive representation is explored.
  \item \textbf{MFLVC} \cite{xu2022Multilevel} proposed a novel contrastive multi-view clustering framework that simultaneously learned low-level and high-level features from multi-view data.
\end{itemize}

\textbf{CSMSC}, \textbf{SM$^2$SC}, \textbf{LMVSC}, \textbf{EOMSC-CA}, and \textbf{CoMSC} are shallow models based on subspace learning. \textbf{MvDSCN} and \textbf{DMSC-UDL} are deep subspace models. \textbf{MCGC} is a graph based model. \textbf{CGL} is a low-rank tensor learning based model. \textbf{MFLVC} is a deep model using contrast strategy.
We run the codes published by the authors and set the parameters according to the interval suggested in these papers.

In the proposed SCMC, we use two kinds of three-layer AEs to encode the multi-view data. The dimensions of each layer of one AE are [$d^{(v)}$, 500], [500, 200], and [200, $c$], respectively. The dimensions of each layer of another AE are [$d^{(v)}$, 200], [200, 100], and [100, $c$], respectively. $d^{(v)}$ and $c$ denote the feature dimension of the $v$-th view's data matrix and the number of clusters. The temperature parameter $\tau$ is fixed to 0.1. $\gamma_{1}$, $\gamma_{2}$ are tuned in $\{500, 1000\}$, $\{0.009, 0.025, 0.03, 0.06,  0.3, 0.4\}$, respectively. $\gamma_{3}$ is set to 0.01.

\subsection{Evaluation Metrics}
Seven dominant clustering evaluation metrics are adopted to quantify the clustering performance, which are Accuracy (ACC), Normalized Mutual Information (NMI), Purity, Adjusted Rand Index (ARI), F-score, Precision, and Recall respectively. In view of the algorithm stability, each experiment is run ten times, then the mean values are reported. The ranges of ACC, NMI, Purity, F-score, Precision, and Recall are all [0, 1], while the range of ARI is [-1, 1]. For all metrics, higher values correspond to  better clustering effects.

\subsection{Experimental Results}
Tables \ref{multiview-clustering-datasets-part1} and \ref{multiview-clustering-datasets-part2} provide the numerical results of all experiments, the best results are bolded and the second best results are underlined.
These results reveal several interesting phenomenons and they are explained below.
From a holistic perspective, the proposed SCMC outperforms other single-view and multi-view clustering methods, which show that the strategies adopted by the model to improve the representation learning ability is efficient.
Especially, on the UCI dataset, SCMC raises the clustering performance by 9\%, 2.28\%, 8.45\%, 9.7\%, 8.68\%, 14.85\%, 1.47\%  compared to the suboptimal results.
MFLVC also leverages the contrast mechanism to improve the information capability of different representations, whose clustering performance achieves favorable status on several datasets such as GRAZ02, Youtube. While SCMC still achieves better performance, this situation may be attributed to the fact that SCMC applies the subspace learning technology to standardizes data from various modalities into a common semantic space, thus alleviating the difficulties in the contrast process.
MvDSCN and DMSC-UDL are two deep multi-view subspace clustering models, from all experimental statistics, their clustering outcomes are not stable. For example, their clustering effects on UCI dataset are relatively superior, and both exceed the baseline method K-means. Nonetheless, on WikipediaArticles dataset, they both produce inferior clustering results than K-means.
An interesting observation is that although CGL is a conventional shallow model, it yields promising clustering results, probably because it captures the high-order correlations between views by virtue of the low-rank tensor learning.
\begin{table*}[htbp]
	\centering
	\setlength{\tabcolsep}{2mm}{
		\caption{Comparison of clustering results (\%) on four real-world datasets.}
		\vspace{0pt}
		\centering
		\begin{tabular}{|c||c||c|c|c|c|c|c|c|c|c|}
			\toprule
             Datasets & Methods & ACC & NMI &Purity & ARI &F-score & Precision & Recall  \\
			\midrule
            \midrule
          \multirow{12}{*}{ALOI}
          & K-means & 47.49   & 47.34 & 48.58   & 32.98  & 41.04   & 33.97 & 52.08
            \\
           & CSMSC & 75.66  & 73.32  & 76.68  & 63.61  & 67.42  & 63.80 & 71.49
            \\
           & MCGC & 83.32  & 74.77  & 83.32  & 65.51  & 69.06  & 66.61 & 71.69
            \\
           & SM$^2$SC & 23.73   & 26.34 & 28.64  & 13.86  & 25.29  & 18.90 & 38.20 \\
           & MvDSCN & 82.39   & 82.08 & 74.88  & 74.56  & 79.56  & 77.87 & 80.67 \\
           & LMVSC & 67.19   & 67.91  & 67.47  & 57.97  & 62.67 & 55.55 & 71.88
            \\
           & CGL  & \underline{94.89}   & \underline{91.54} & \underline{94.89}  & \underline{89.18}  & \underline{90.25}  & \underline{90.03} & \underline{90.47} \\
           & DMSC-UDL & 55.79   & 52.99 & 59.68  & 38.68  & 48.88  &  45.52 & 52.78
            \\
           & EOMSC-CA & 65.80   & 76.36 & 66.27   & 47.24 &  54.49  &  39.33 & 88.70
            \\
           & CoMSC & 80.63   & 81.31  & 84.62   & 71.96 & 74.95  &  69.63 & 81.15
            \\
           & MFLVC & 82.63   & 78.57  & 72.85   & 69.59 & 73.89  &  73.84 & 73.94
            \\
           & SCMC & \textbf{95.74}   & \textbf{92.12} & \textbf{95.74}  & \textbf{90.91}  & \textbf{92.16}  & \textbf{92.18} &\textbf{92.14}\\
           \midrule
           \midrule
           \multirow{12}{*}{GRAZ02}
          & K-means & 35.91   & 3.20 & 35.91  & 3.56  & 33.83   & 27.19 & \textbf{44.78}           \\
           & CSMSC & -  & - & -  & -  & -  & - & -
            \\
            & MCGC & 42.75  & 6.88  & 42.75  & 6.38  & 30.09  & 29.98 & 30.91  \\
            & SM$^2$SC & 47.15   & 12.49 & 47.76  & 11.91  & 34.22  & 34.09 & 34.35 \\
            & MvDSCN & 40.44   & 6.81 & 50.54  & 5.93  & 31.24  & 29.97 & 32.62 \\
            & LMVSC & 44.24   & 8.21 & 44.24 & 8.09  & 31.40  & 31.23 & 31.57
            \\
            & CGL & 46.46   & 12.54 & 46.46  & 11.43  & 33.87  & 33.72 & 34.03
            \\
            & DMSC-UDL & 41.32   & 8.33   & 52.24 & 8.28   & 32.65  &  31.05 & 34.42
            \\
            & EOMSC-CA & 42.48   & 12.19  & 46.95   & 10.66 &  33.33  &  33.14 & 33.51
            \\
           & CoMSC & 40.79   & 8.29  & 43.50   & 7.82 &  31.17  &  31.04 & 31.30
            \\
            & MFLVC & \underline{47.97}   & \underline{13.76}  & \underline{57.18}  & \underline{13.79} & \underline{35.64}  & \underline{35.35} & 35.93
            \\
           & SCMC & \textbf{51.90}   & \textbf{16.16} & \textbf{59.55}  & \textbf{14.11 } & \textbf{37.34}  & \textbf{37.72} &\underline{36.97}
            \\
            \midrule
            \midrule
            \multirow{12}{*}{NUS-WIDE-v1}
           & K-means & 25.44  & 14.88  & 28.69 & 6.79  & 23.71   & 16.26 & \underline{43.73}
            \\
           & CSMSC & 34.31  & 19.58 & \underline{38.38} & \underline{13.29}  & 24.46  & 23.63 & 25.34
            \\
            & MCGC & 31.75  & 18.87  & 35.63  & 11.35  & 23.40  & 21.42 & 25.78  \\
            & SM$^2$SC & 32.31   & 19.87 & 36.81  & 12.66  & 23.83  & 23.18 & 24.52 \\
            & MvDSCN & 31.13   & 16.76 & 35.19  & 10.01  & 23.66  & 23.72 & 23.59 \\
           & LMVSC & 30.19 & 19.29  & 36.00  & 11.29  & 23.19 & 21.51 & 25.17            \\
            & CGL & 31.35  & \underline{20.36} & 36.91   &  12.57  &  24.52 & 22.31 & 27.22
            \\
            & DMSC-UDL & 25.44   & 9.37  & 30.69 & 5.86    & 20.04  &  17.26 & 23.89
            \\
            & EOMSC-CA & 32.94   & 20.21  & 34.00  & 11.84 &  \underline{27.11}  &  19.48 & \textbf{44.55}
            \\
           & CoMSC & 28.63   & 11.82  & 29.88   & 9.27 &  20.77  &  20.40 & 21.14
            \\
           & MFLVC & \underline{34.81}   & 18.68  & 36.13   & \textbf{14.33} & 26.55  &  \underline{24.62} & 28.81
            \\
           & SCMC & \textbf{36.56}   & \textbf{21.83} & \textbf{40.06}  & \textbf{14.33} & \textbf{27.92}  & \textbf{28.15} &27.69
            \\
            \midrule
            \midrule
            \multirow{12}{*}{NUS-WIDE-v2}
           & K-means & 15.15   & \underline{19.62}  & 25.55 & 4.91  & 9.88   & 11.16 & 8.86
            \\
           & CSMSC & 13.60  & 15.89  & 21.90  & 3.08 & 7.55  & 9.68 & 6.18
            \\
            & MCGC & 15.05  & 15.92  & 21.60  & 3.27  & 9.20  & 8.87 & 9.56  \\
            & SM$^2$SC & 15.15   & 17.16 & 24.75  & 4.46  & 8.49  & 11.89 & 6.60\\
            & MvDSCN & 14.55   & 17.90 & \underline{25.80}  & 3.71  & 10.05  & \underline{12.87} & 8.25 \\
           & LMVSC & 15.40   & 18.13  & 24.35  & 4.66  & 8.76 & 12.04 & 6.88
            \\
            & CGL & 14.37  & 17.53  & 23.85  & 3.79 &  7.96  & 10.86 & 6.27
            \\
            & DMSC-UDL & 15.85   & 14.55 & 20.35 & 1.69  & \textbf{14.26}  &  10.84 & \underline{20.83}
            \\
            & EOMSC-CA & 15.05   & 13.33  & 18.90   & 3.92 &  11.17  &  6.10 & \textbf{65.71}
            \\
           & CoMSC & 12.85   & 14.39  & 20.60   & 2.39 & 6.69  &  8.98 & 5.33
            \\
            & MFLVC & \underline{16.35}   & 14.16  & 23.15   & \underline{5.35} & \underline{13.31}  &  10.45 & 18.31
            \\
           & SCMC & \textbf{17.85}   & \textbf{21.23} & \textbf{30.30}  & \textbf{5.92} & 11.97  & \textbf{15.10} & 9.92
            \\
            \midrule
            \bottomrule
		\end{tabular}
		\label{multiview-clustering-datasets-part1}}
\end{table*}
\begin{figure*}[htbp]
	\centering
	\includegraphics[width=1\textwidth]{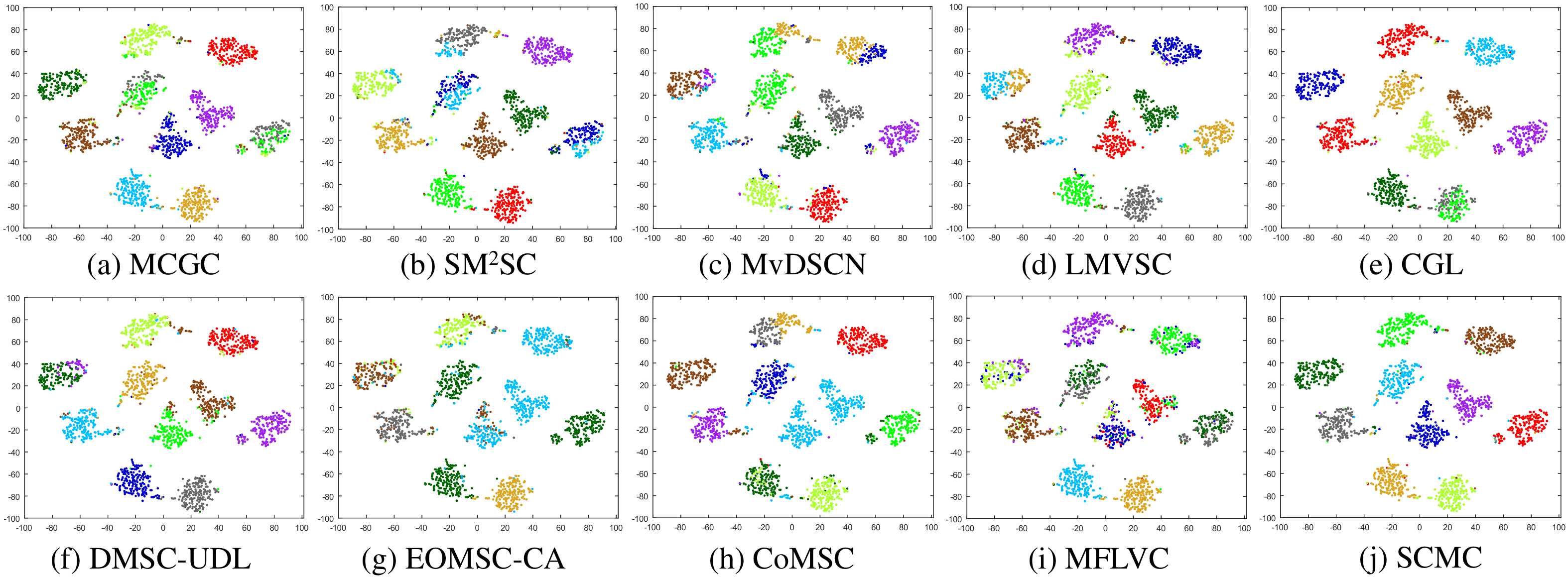}
	\caption{Visualization of clustering results of ten multi-view  clustering methods on UCI dataset via t-SNE technology.}
	\label{sandian_diagram}
\end{figure*}

\begin{table*}[htbp]
	\centering
	\setlength{\tabcolsep}{2mm}{
		\caption{Comparison of clustering results (\%) on four real-world datasets.}
		\vspace{0pt}
		\centering
		\begin{tabular}{|c||c||c|c|c|c|c|c|c|}
			\toprule
             Datasets & Methods & ACC & NMI &Purity & ARI &F-score & Precision & Recall \\
			\midrule
            \midrule
            \multirow{12}{*}{Reuters}
           & K-means & 42.47  & 22.09  & 50.67 & 16.19  & 36.63   & 32.40 &42.14
            \\
           & CSMSC & \underline{50.27}  & \underline{30.12}  & \underline{58.67}  & \textbf{24.83}  & 42.06  & 39.58 & 44.86
            \\
            & MCGC & 35.33  & 10.15  & 42.47  & 4.15  & 34.09  & 23.54 & \underline{61.77}  \\
            & SM$^2$SC & 47.93  & 26.06  & 52.40  & \underline{24.76}  & 41.25  & 40.59 & 41.93
            \\
            & MvDSCN & 49.20   & 28.59 & 50.67  & 19.71  & \underline{42.46}  & 41.46 & 43.52 \\
           & LMVSC & 47.40  & 26.80 & 51.73  & 19.64  & 41.27 & 33.07 & 54.88
            \\
            & CGL & 44.59 & 21.21  & 48.97   & 20.83 &  37.35 &  38.61 & 36.17            \\
            & DMSC-UDL & 31.60   & 9.42 & 41.13  & -0.67   & 40.24  &  26.58 & \textbf{82.75}
            \\
            & EOMSC-CA & 37.60  & 12.53  & 46.40   & 12.32 &  30.20  &  31.95 & 28.63
            \\
           & CoMSC & -   & -  & -   & - &  -  &  - & -
            \\
            & MFLVC & 43.40   & 29.76  & \textbf{60.53}   & 24.70 & 38.13  &  \underline{44.23} & 33.51
            \\
           & SCMC & \textbf{51.80}   & \textbf{34.47} & 53.87  & 21.83 & \textbf{50.97}  & \textbf{44.93} & 58.88
            \\
            \midrule
            \midrule
            \multirow{12}{*}{UCI}
           & K-means & 38.76   & 46.64  & 44.23 & 31.35  & 38.86   & 35.39 & 43.10
            \\
           & CSMSC & 78.75  & 76.97  & 81.30  & 70.75  & 73.74  & 71.79 & 75.57
            \\
            & MCGC & 80.20  & 79.74  & 83.55  & 73.90  & 76.63  & 72.87 & 80.80  \\
            & SM$^2$SC & 84.20   & 79.94 & 84.20  & 74.86  & 77.37  & 77.18 & 77.56 \\
            & MvDSCN & 81.85   & 71.72 & 73.05  & 65.82  & 69.47  & 69.31 & 69.64 \\
           & LMVSC & \underline{87.75}  & 80.55  & 87.75  & 76.58  & 78.93 & 78.27 & 79.59
            \\
            & CGL & 84.25   & \underline{90.52}  & \underline{88.55}  & \underline{83.22} &  \underline{85.02}  &  \underline{78.86} & \underline{92.23}
            \\
            & DMSC-UDL & 75.80   &72.62 & 79.55  & 63.68   & 70.96  &  69.58 & 72.64
            \\
            & EOMSC-CA & 54.80   & 67.09  & 55.10   & 46.28 &  53.57 &  39.29 & 84.17
            \\
            & CoMSC & 77.80 & 78.41 & 81.55   & 69.28 &  72.60  &  66.85 & 79.44
            \\
            & MFLVC & 79.95   & 78.36  & 79.95   & 69.73 & 73.31  &  72.51 & 74.12
            \\
           & SCMC & \textbf{96.75}   & \textbf{92.84}  & \textbf{97.00}  & \textbf{92.92} & \textbf{93.70}  & \textbf{93.71} & \textbf{93.70}
            \\
            \midrule
            \midrule

            \multirow{12}{*}{WikipediaArticles}
           & K-means & 54.69   & 51.48  & 58.59 & 39.02 & 45.80   & 44.97 & 46.67
            \\
           & CSMSC & 52.03  & 46.47  & 55.70  & 38.36  & 44.91  & 46.16 & 43.72
            \\
            & MCGC & 54.40  & 40.79  & 56.85  & 31.68  & 38.91  & 40.13 & 37.77  \\
            & SM$^2$SC & 55.12  & 50.83  & 59.31  & 40.67  & 46.95  & 48.46 & 45.53
            \\
            & MvDSCN & 34.63   & 29.74 & 41.85  & 17.58  & 29.79  & 28.14 & 31.65 \\
           & LMVSC & 55.56 & 47.46 & 57.00  & 33.12  & 41.00 & 37.99 &44.53
            \\
            & CGL & 54.16   & 49.83  & 59.42  & 37.09 &  44.11  &  43.21 & 45.05
            \\
            & DMSC-UDL & 38.24   &31.16 & 18.62  & \textbf{45.02}   & 33.09  &  31.47 & 34.89
            \\
            & EOMSC-CA & \underline{56.13}   & \underline{52.91}  & \underline{61.04}   & \underline{42.30} &  \underline{48.47} &  \underline{49.56} &  \underline{47.42}
            \\
           & CoMSC & 21.07  & 7.18  & 23.23   & 2.90 &  13.13  &  13.61 & 12.68
            \\
            & MFLVC & 40.40   & 31.63  & 43.00 & 22.57 & 34.06  &  26.99 & 46.18
            \\
           & SCMC & \textbf{58.30}   & \textbf{53.86} & \textbf{66.96}  & 37.52 & \textbf{49.79}  & \textbf{51.94} & \textbf{47.82}
            \\
            \midrule
            \multirow{12}{*}{Youtube}
           & K-means & 24.56   & 15.16  & 27.85 & 8.19  & 19.48  & 15.78 & 25.55
            \\
           & CSMSC & -   & -    & -& -  & -   & - & -
            \\
            & MCGC & 28.35  & 13.66  & 29.70  & 8.34  & 17.62  & 17.33 & 17.93  \\
            & SM$^2$SC & 30.58  & 18.12  & 34.08  & 11.14  & 20.11  & 19.83 & 20.40
            \\
            & MvDSCN & 30.55   & 17.67 & 34.90  & 10.76  & 22.77  & 20.41 & 25.75 \\
            & LMVSC &  27.20 & 14.79  & 29.15  & 7.93  & 17.32 & 16.91 & 17.74
            \\
           & CGL & \underline{32.95}   & 19.65  & 35.39  & 12.52 &  21.48  &  20.88 & 22.11 \\
           & DMSC-UDL & -   & - & -  & -   & -  &  - & -
            \\
            & EOMSC-CA & 32.20   & 18.17  & 32.60   & 12.82 &  23.49  &  19.11 & 30.45
            \\
           & CoMSC & 24.60   & 10.41  & 26.00   & 5.87 &  15.42  &  15.13 & 15.71
            \\
            & MFLVC & 31.80   & \underline{24.37}  & \underline{38.00}   & \underline{15.98} & \underline{26.01}  &  \underline{22.22} & \underline{31.36}
            \\
           & SCMC & \textbf{37.90}   & \textbf{26.12}  & \textbf{41.15}  & \textbf{18.53} & \textbf{28.50}  & \textbf{25.61} & \textbf{32.14}
            \\
            \midrule
            \bottomrule
		\end{tabular}
		\label{multiview-clustering-datasets-part2}}
\end{table*}

\subsection{Experimental Visualization}
We undertake some visualization experiments to directly compare the divergence in representation learning abilities of varying approaches.
The t-SNE technology is used to downscale the concatenated UCI dataset to a two-dimensional planes.
Fig. \ref{sandian_diagram} visualizes the clustering results of ten multi-view clustering methods on the UCI dataset, and data points belonging to different clusters are painted in different colors.
It can be seen that the ten clusters divided by SCMC rarely involve the samples of other clusters,  the segmentation is nearly perfect.
On the contrary, the clusters produced by EOMSC-CA are heavily mixed with samples of the other clusters, and even cannot divide enough ten clusters.

\begin{figure*}[htbp]
	\centering
	\includegraphics[width=1\textwidth]{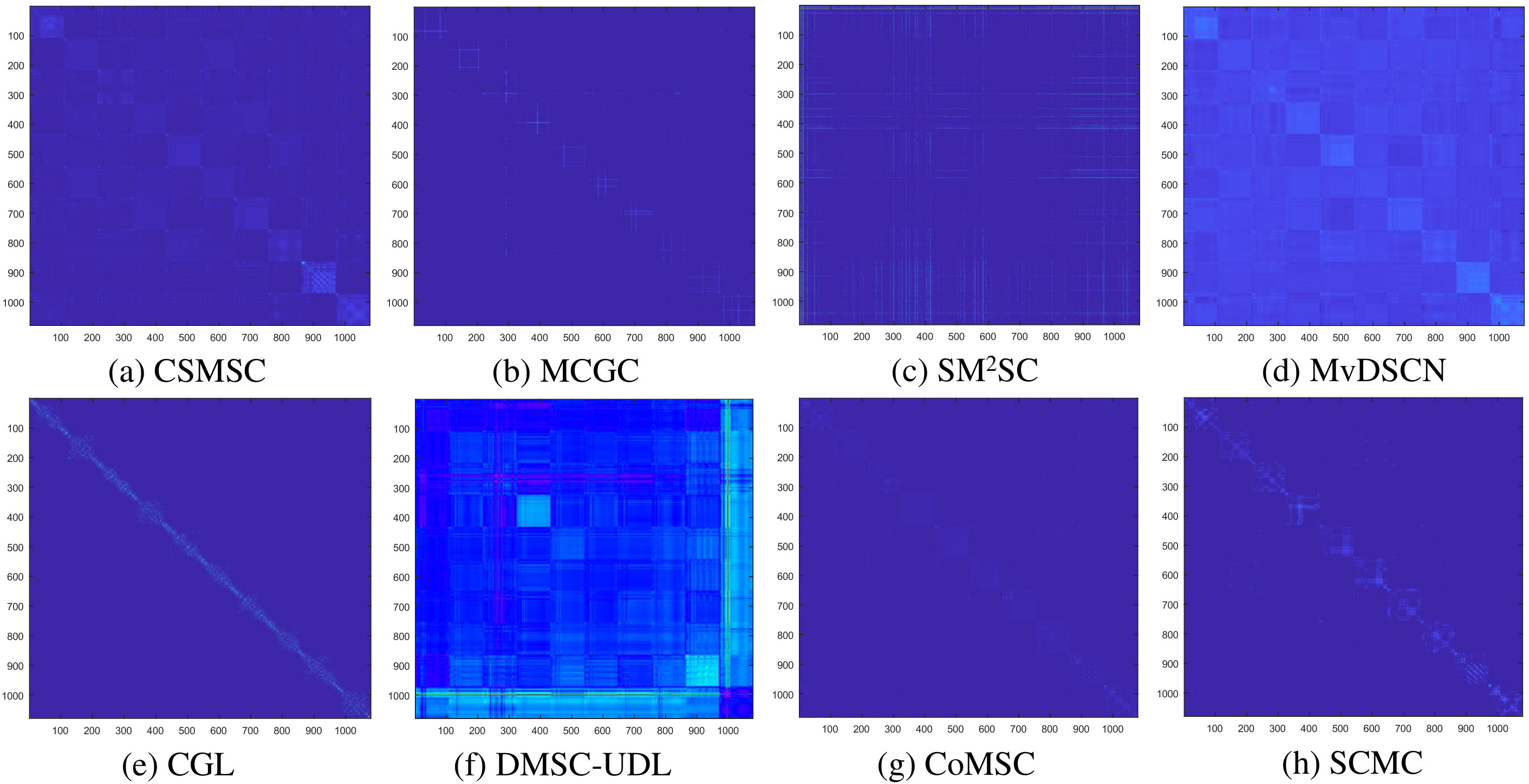}
	\caption{Visualizations of consistent affinity representations learned via eight multi-view clustering approaches on ALOI dataset.}
	\label{yingxiang_diagram}
\end{figure*}

\begin{figure*}[htbp]
	\centering
	\includegraphics[width=0.8\textwidth]{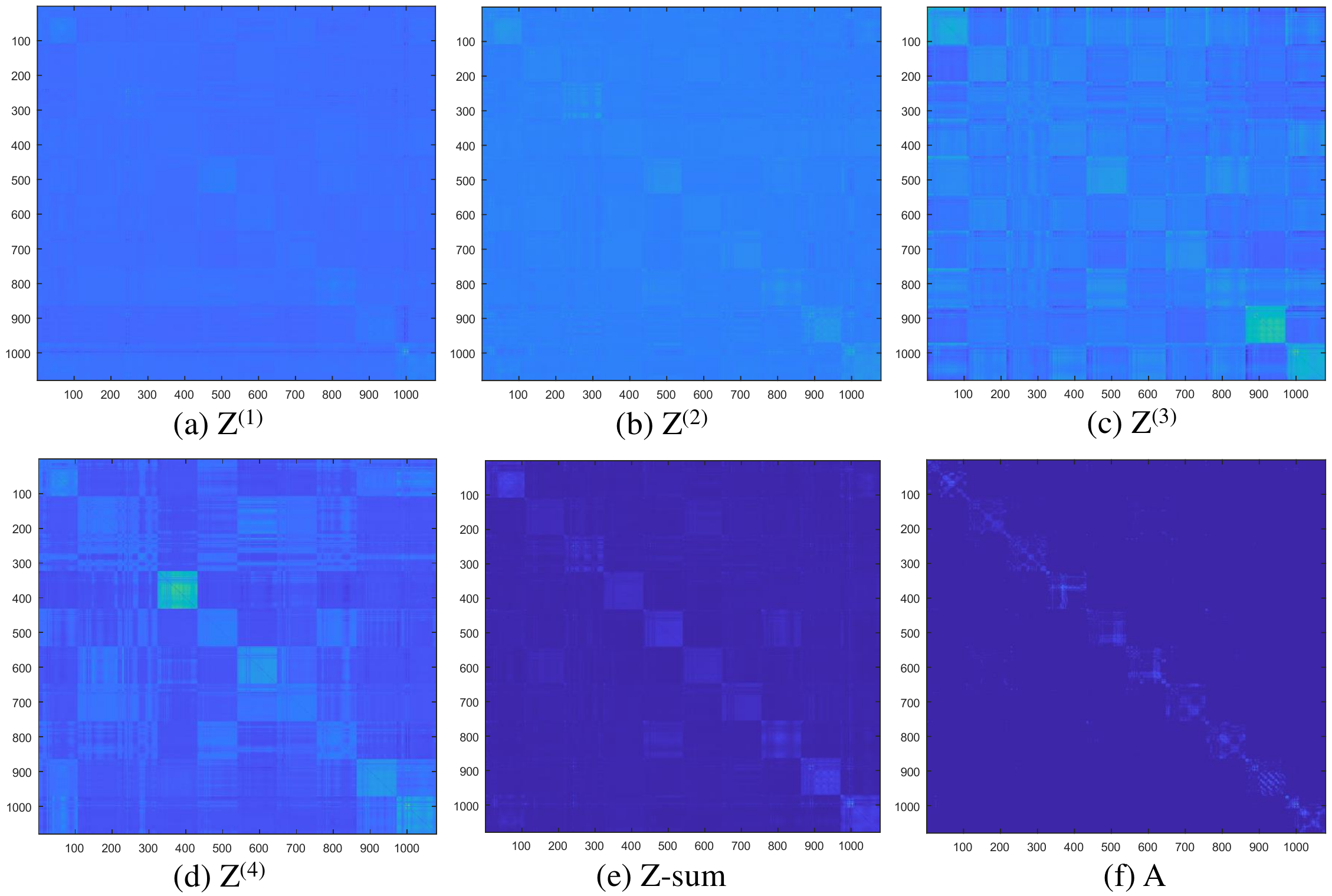}
	\caption{Comparison of view-specific and consistent affinity matrices generated by the proposed SCMC on ALOI dataset. }
	\label{yingxiang_ALOI}
\end{figure*}

Fig. \ref{yingxiang_diagram} presents the visualizations of consistent affinity representations learned by seven compared multi-view clustering methods and the proposed SCMC. We can observe that SM$^2$SC and DMSC-UDL
almost fail to engrave the diagonal-block structures, CGL can clearly highlight important structures on the
diagonal, but the boundaries between diagonal-blocks are not apparent.
Fortunately, SCMC achieves the clear depiction of diagonal-blocks. In addition, the view-specific and consistent affinity representations produced by SCMC are also visualized in Fig. \ref{yingxiang_ALOI}.
As Fig. \ref{yingxiang_ALOI} shows, the capacity of the affinity matrices of four single views to characterize the similarities between instances is not good, Fig. \ref{yingxiang_ALOI} (e) is generated via averaging all subspace representations, namely $\mathbf{Z}$-sum  $=(\mathbf{Z}^{(1)}+\mathbf{Z}^{(2)}+\mathbf{Z}^{(3)}+\mathbf{Z}^{(4)})/4$.
Though $\mathbf{Z}$-sum can also clearly portray the diagonal-blocks structures, its suppression of non-diagonal affinities is weak, which means that it does not protect the local geometric structures of data well.

\begin{figure*}[htbp]
	\centering
	\includegraphics[width=1\textwidth]{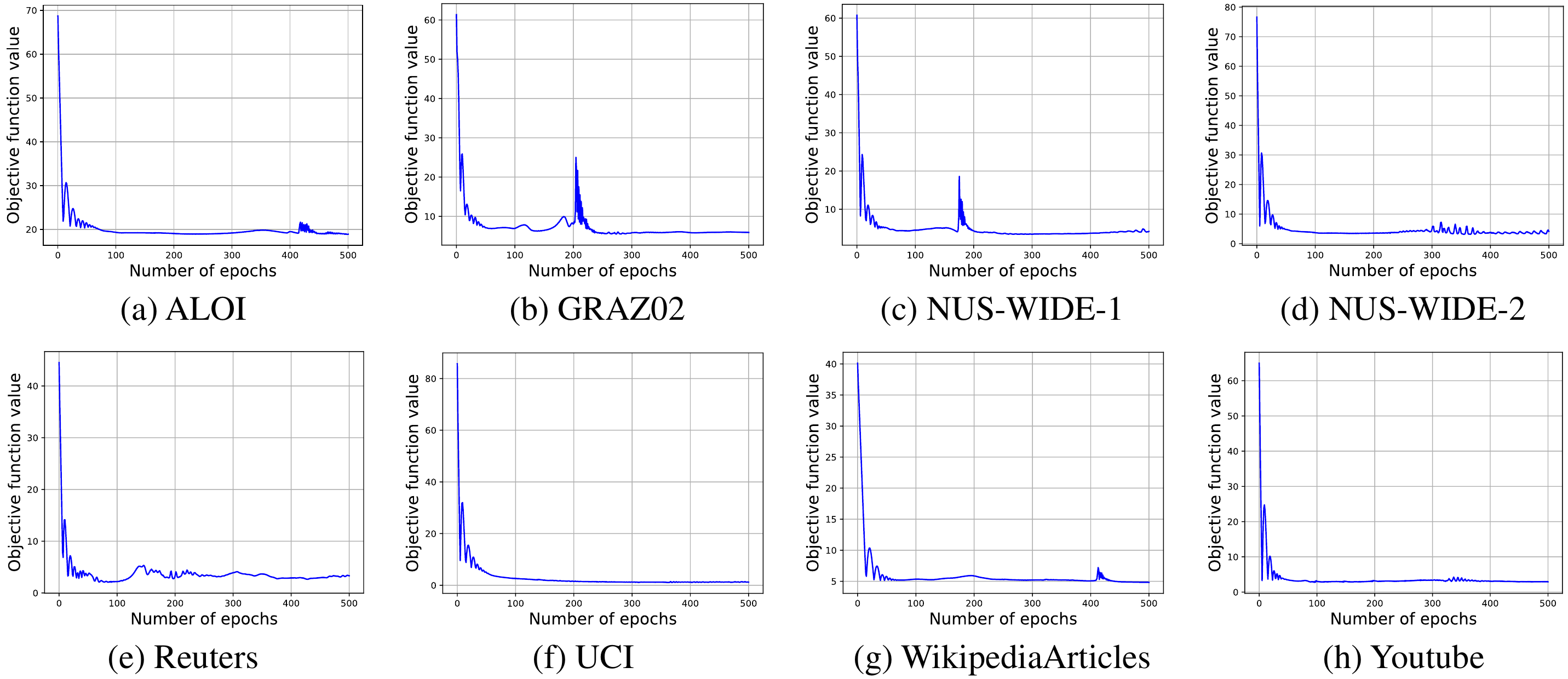}
	\caption{The convergence curves of the proposed SCMC on eight datasets. }
	\label{convergence-diagram}
\end{figure*}

To evaluate the convergence of the proposed SCMC, we plot the curves of the objective function values as the number of training epochs increase on eight datasets in Fig. \ref{convergence-diagram}. It can be observed that the objective function values
decrease quickly, though there are fluctuations in the middle of training process, they eventually tend to stabilize.

\subsection{Ablation Study}
The proposed SCMC consist of four loss components, i.e., $\mathcal{L}_{Re}$, $\mathcal{L}_{Sub}$, $\mathcal{L}_{Con}$, and $\mathcal{L}_{Fu}$.
We implement experiments to verify the role played by each loss component.
Tables \ref{ablation1} and \ref{ablation2} report the ablation results.
When SCMC has only reconstruction loss $\mathcal{L}_{Re}$, we obtain the consensus features by averaging multiple embedding features, then they are fed into the K-means algorithm to yield the clustering results. However, the results are inferior,  which means that the latent representations embedded only through AEs are not yet well discriminated.
When the subspace loss is introduced, i.e., $\mathcal{L}_{Re}+\mathcal{L}_{Sub}$, the performance is somewhat improved.
It is worth noting that once the contrast loss is included, the improvements of clustering effects are significant, illustrating that the contrast strategy dose help to augment the discrimination of subspace representations.
In general, on the basis of $\mathcal{L}_{Re}+\mathcal{L}_{Sub}$, SCMC performs better with the fusion loss $\mathcal{L}_{Fu}$ than with contrast loss $\mathcal{L}_{Con}$, which indicates that the contribution of $\mathcal{L}_{Fu}$ is greater. Nevertheless, $\mathcal{L}_{Con}$ is essential if the optimal performance is to be achieved.
\begin{table*}[]
\centering
	\setlength{\tabcolsep}{2mm}{
		\caption{Ablation results of the proposed SCMC on four datasets.}
		\vspace{0pt}
		\centering
\begin{tabular}{|c|ccc|ccc|ccc|ccc|}
\toprule
\multicolumn{1}{|c|}{\multirow{2}{*}{Loss}} & \multicolumn{3}{c|}{ALOI} & \multicolumn{3}{c|}{GRAZ02} & \multicolumn{3}{c|}{NUS-WIDE-v1} & \multicolumn{3}{c|}{NUS-WIDE-v2}  \\ \cline{2-13}
\multicolumn{1}{|c|}{}   & ACC    & NMI    & Purity    & ACC    & NMI    & Purity  & ACC    & NMI    & Purity    & ACC    & NMI    & Purity\\
\midrule
$\mathcal{L}_{Re}$                     & 71.36 --         & 82.16 --           & 76.09 --           & 38.28 -- & 7.30 -- & 54.47 --  & 27.00 -- & 12.05 -- & 35.06 --  & 14.40 -- & 18.29 -- & 25.85 -- \\
$\mathcal{L}_{Re}$+$\mathcal{L}_{Sub}$ & 78.68 $\uparrow$ & 81.16 $\downarrow$ & 72.47 $\downarrow$ & 39.77 $\uparrow$ & 8.74 $\uparrow$ & 55.08 $\uparrow$ & 30.69 $\uparrow$ & 19.82 $\uparrow$ & 36.18 $\uparrow$ & 15.35 $\uparrow$ & 16.58 $\downarrow$ & 27.40 $\uparrow$\\
$\mathcal{L}_{Re}$+$\mathcal{L}_{Sub}$+$\mathcal{L}_{Con}$  & 88.32 $\uparrow$ & 84.17 $\uparrow$ & 79.98 $\uparrow$ & 45.33 $\uparrow$ & 12.46 $\uparrow$ & 56.23 $\uparrow$ & 32.50 $\uparrow$ & 19.88 $\uparrow$ & 36.88 $\uparrow$ & 15.80 $\uparrow$ & 17.59 $\downarrow$ & 25.75 $\downarrow$\\
$\mathcal{L}_{Re}$+$\mathcal{L}_{Sub}$+$\mathcal{L}_{Fu}$  & 93.05 $\uparrow$ & 89.87 $\uparrow$ & 84.52 $\uparrow$ & 45.12 $\uparrow$ & 11.18 $\uparrow$ & 57.18 $\uparrow$ & 32.88 $\uparrow$ & 18.82 $\uparrow$ & 36.88 $\uparrow$ & 15.70 $\uparrow$ & 19.03 $\uparrow$ & 27.80 $\uparrow$\\
\midrule
$\mathcal{L}$         & \textbf{95.74} $\uparrow$   & \textbf{92.72} $\uparrow$  & \textbf{95.74} $\uparrow$ & \textbf{51.90} $\uparrow$  & \textbf{16.16} $\uparrow$ & \textbf{59.55} $\uparrow$ & \textbf{36.56} $\uparrow$ & \textbf{21.83} $\uparrow$ & \textbf{40.06} $\uparrow$ & \textbf{17.85} $\uparrow$ & \textbf{21.23} $\uparrow$ & \textbf{30.30} $\uparrow$\\
\bottomrule
\end{tabular}
\label{ablation1}}
\end{table*}

\begin{table*}[]
\centering
	\setlength{\tabcolsep}{2mm}{
		\caption{Ablation results of the proposed SCMC on four datasets.}
		\vspace{0pt}
		\centering
\begin{tabular}{|c|ccc|ccc|ccc|ccc|}
\toprule
\multicolumn{1}{|c|}{\multirow{2}{*}{Loss}} & \multicolumn{3}{c|}{Reuters} & \multicolumn{3}{c|}{UCI} & \multicolumn{3}{c|}{WikipediaArticles} & \multicolumn{3}{c|}{Youtube}  \\ \cline{2-13}
\multicolumn{1}{|c|}{}   & ACC    & NMI    & Purity    & ACC    & NMI    & Purity  & ACC    & NMI    & Purity    & ACC    & NMI    & Purity\\
\midrule
$\mathcal{L}_{Re}$ & 41.80 -- & 17.53 --  & 45.93 --   & 70.90 -- & 69.16 -- & 73.90 --  & 23.09 -- & 9.59 -- & 29.73 -- & 23.95 -- & 13.00 -- & 29.15 -- \\
$\mathcal{L}_{Re}$+$\mathcal{L}_{Sub}$ & 38.33 $\downarrow$ & 28.52 $\uparrow$ & 51.07 $\uparrow$ & 75.65 $\uparrow$ & 73.88 $\uparrow$ & 78.65 $\uparrow$ & 35.64 $\uparrow$ & 24.54 $\uparrow$ & 40.69 $\uparrow$ & 27.65 $\uparrow$ & 17.65 $\uparrow$ & 32.40 $\uparrow$\\
$\mathcal{L}_{Re}$+$\mathcal{L}_{Sub}$+$\mathcal{L}_{Con}$  & 47.60 $\uparrow$ & 31.05 $\uparrow$ & 49.93 $\uparrow$ & 91.40 $\uparrow$ & 83.70 $\uparrow$ & 81.90 $\uparrow$ & 37.95 $\uparrow$ & 27.72 $\uparrow$ & 41.99 $\uparrow$ & 29.20 $\uparrow$ & 18.74 $\uparrow$ & 35.40 $\uparrow$\\
$\mathcal{L}_{Re}$+$\mathcal{L}_{Sub}$+$\mathcal{L}_{Fu}$  & 48.00 $\uparrow$ & 29.34 $\uparrow$ & 51.00 $\uparrow$ & 95.55 $\uparrow$ & 90.99 $\uparrow$ & 86.25 $\uparrow$ & 56.57 $\uparrow$ & \textbf{54.67} $\uparrow$ & 64.36 $\uparrow$ & 31.90 $\uparrow$ & 18.90 $\uparrow$ & 37.60 $\uparrow$\\
\midrule
$\mathcal{L}$   & \textbf{51.80} $\uparrow$ & \textbf{34.47} $\uparrow$  & \textbf{53.87} $\uparrow$  & \textbf{96.75} $\uparrow$  & \textbf{92.85} $\uparrow$ & \textbf{97.00} $\uparrow$  & \textbf{58.30} $\uparrow$ & 53.86 $\uparrow$ & \textbf{66.96} $\uparrow$  & \textbf{37.90} $\uparrow$ & \textbf{26.12} $\uparrow$ & \textbf{41.15} $\uparrow$\\
\bottomrule
\end{tabular}
\label{ablation2}}
\end{table*}
\subsection{Parameter Sensitivity Analysis}
\begin{figure*}[htbp]
	\centering
	\includegraphics[width=1\textwidth]{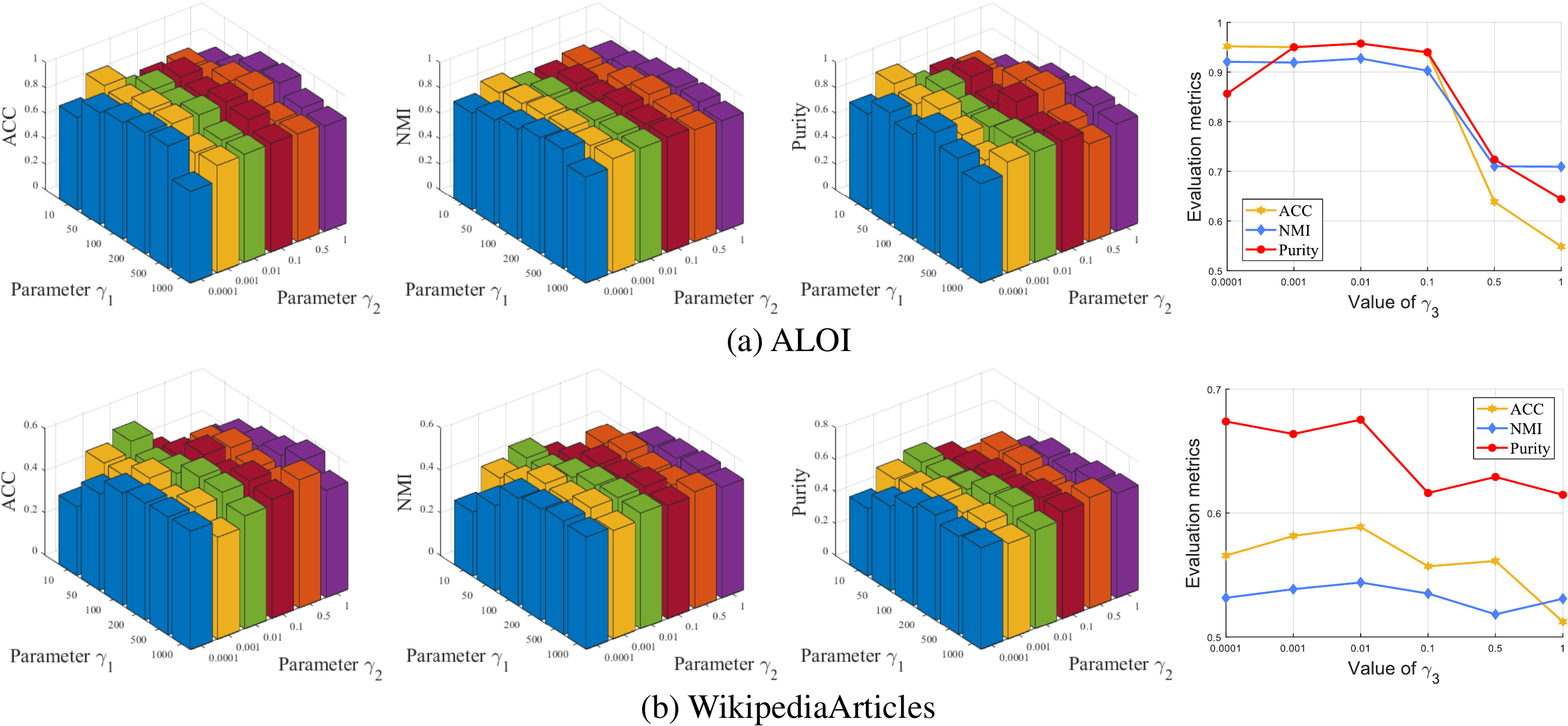}
	\caption{The parameter sensitivity analysis of the proposed SCMC on ALOI and WikipediaArticles datasets. }
	\label{para_analyse_IEEE}
\end{figure*}

Three nonnegative trade-off parameters are used to balance the overall objective loss, their impacts on the clustering outcomes are investigated via sensitivity experiments.
Specifically, we tune $\gamma_{1}$, $\gamma_{2}$, and $\gamma_{3}$ in \{10, 50, 100, 200, 500, 1000\}, \{0.0001, 0.001, 0.01, 0.1, 0.5, 1\}, and \{0.0001, 0.001, 0.01, 0.1, 0.5, 1\}, respectively.
Fig. \ref{para_analyse_IEEE} shows the numerical results of SCMC under different settings of $\gamma_{1}$, $\gamma_{2}$, and $\gamma_{3}$ . When evaluating $\gamma_{1}$ and $\gamma_{2}$, we fix $\gamma_{3}$, the same operation is done for evaluating $\gamma_{3}$.
It can been seen that a small $\gamma_{1}$ is not conducive to learning an informative subspace representation, this is caused by the insufficient penalty for $||\mathbf{C}^{(v)^{T}}-\mathbf{C}^{(v)^{T}}\mathbf{Z}^{(v)}||_{F}^{2}$ in the optimization. Similarly, when $\gamma_{2}$ is set to be small, the clustering performance is not ideal, because the contrast mechanism has little effects to enhance the discrimination of subspace representations. Interestingly, selecting a relatively lager $\gamma_{3}$ can degrade the effectiveness of SCMC, this situation may result from over-smoothing.

\section{Conclusion}\label{conclusion}
In this paper, we propose a novel Subspace-Contrastive Multi-View Clustering approach (SCMC).
In SCMC, a set of view-specific auto-encoders are adopted to condense the multi-view data and capture its nonlinear structures, thus obtaining the compact data features.
In the new feature spaces, the subspace representation of each view is explored through a self-expression layer. Inspired by the idea of contrastive learning, we regard the subspace representation of each view as a contrastable entity.
By pairwise comparison of multiple subspace representations, we exploit the complementary information in them and augment the discriminability of each subspace representation.
Guided by the important principle of multi-view consensus, we obtain a consistent affinity matrix by combining all subspace representations with a weighted fusion scheme. Furthermore, the graph regularization is employed to encode the local geometric structures embedded in multiple subspaces.
Although the contrast manner used in this paper can upgrade the subspace representation learning, this instance-level contrast manner may pull the samples belonging to the same clusters far apart. Therefore, in future work, we will investigate an adaptive cluster-level contrast paradigm to dissipate this potential effect.

\bibliographystyle{ieeetr}
\bibliography{DCMSC-reference}

\begin{thebibliography}{10}

\bibitem{FurecentMulti2020}
L.~Fu, P.~Lin, A.~V. Vasilakos, and S.~Wang, ``An overview of recent multi-view
  clustering,'' {\em Neurocomputing}, vol.~402, pp.~148--161, 2020.

\bibitem{zhao2017multilearning}
J.~Zhao, X.~Xie, X.~Xu, and S.~Sun, ``Multi-view learning overview: Recent
  progress and new challenges,'' {\em Information Fusion}, vol.~38, pp.~43--54,
  2017.

\bibitem{zong2020Multimodal}
L.~Zong, F.~Miao, X.~Zhang, and B.~Xu, ``Multimodal clustering via deep
  commonness and uniqueness mining,'' in {\em Proceedings of the {ACM}
  International Conference on Information and Knowledge Management},
  pp.~2357--2360, 2020.

\bibitem{wang2020GMC}
H.~{Wang}, Y.~{Yang}, and B.~{Liu}, ``Gmc: Graph-based multi-view clustering,''
  {\em IEEE Transactions on Knowledge and Data Engineering}, vol.~32, no.~6,
  pp.~1116--1129, 2020.

\bibitem{wen2021Adaptive}
J.~Wen, K.~Yan, Z.~Zhang, Y.~Xu, J.~Wang, L.~Fei, and B.~Zhang, ``Adaptive
  graph completion based incomplete multi-view clustering,'' {\em {IEEE}
  Transactions on Multimedia}, vol.~23, pp.~2493--2504, 2021.

\bibitem{wen2021Structural}
J.~Wen, Z.~Wu, Z.~Zhang, L.~Fei, B.~Zhang, and Y.~Xu, ``Structural deep
  incomplete multi-view clustering network,'' in {\em Proceedings of the {ACM}
  International Conference on Information and Knowledge Management},
  pp.~3538--3542, 2021.

\bibitem{Li2021Consensus}
Z.~Li, C.~Tang, X.~Liu, X.~Zheng, G.~Yue, W.~Zhang, and E.~Zhu, ``Consensus
  graph learning for multi-view clustering,'' {\em IEEE Transactions on
  Multimedia}, 2021.
\newblock doi={10.1109/TMM.2021.3081930}.

\bibitem{hajjar2022constrained}
S.~E. Hajjar, F.~Dornaika, and F.~Abdallah, ``Multi-view spectral clustering
  via constrained nonnegative embedding,'' {\em Information Fusion}, vol.~78,
  pp.~209--217, 2022.

\bibitem{Hajjar2022spectral}
S.~E. Hajjar, F.~Dornaika, and F.~Abdallah, ``One-step multi-view spectral
  clustering with cluster label correlation graph,'' {\em Information
  Sciences}, vol.~592, pp.~97--111, 2022.

\bibitem{xu2020nonnegativeembedding}
Z.~Hu, F.~Nie, R.~Wang, and X.~Li, ``Multi-view spectral clustering via
  integrating nonnegative embedding and spectral embedding,'' {\em Information
  Fusion}, vol.~55, pp.~251--259, 2020.

\bibitem{yang2021Orthogonal}
B.~Yang, X.~Zhang, F.~Nie, F.~Wang, W.~Yu, and R.~Wang, ``Fast multi-view
  clustering via nonnegative and orthogonal factorization,'' {\em {IEEE}
  Transactions on Image Processing}, vol.~30, pp.~2575--2586, 2021.

\bibitem{huang2022Factorization}
S.~Huang, Y.~Zhang, L.~Fu, and S.~Wang, ``Learnable multi-view matrix
  factorization with graph embedding and flexible loss,'' {\em IEEE
  Transactions on Multimedia}, 2022.
\newblock doi={10.1109/TMM.2022.3157997}.

\bibitem{gao2015Subspace}
H.~Gao, F.~Nie, X.~Li, and H.~Huang, ``Multi-view subspace clustering,'' in
  {\em Proceedings of the IEEE International Conference on Computer Vision},
  pp.~4238--4246, 2015.

\bibitem{zhang2020generalized}
C.~Zhang, H.~Fu, Q.~Hu, X.~Cao, Y.~Xie, D.~Tao, and D.~Xu, ``Generalized latent
  multi-view subspace clustering,'' {\em IEEE Transactions on Pattern Analysis
  and Machine Intelligence}, vol.~42, no.~1, pp.~86--99, 2020.

\bibitem{li2022Correlation}
Z.~Li, C.~Tang, X.~Zheng, X.~Liu, W.~Zhang, and E.~Zhu, ``High-order
  correlation preserved incomplete multi-view subspace clustering,'' {\em
  {IEEE} Transactions on Image Processing}, vol.~31, pp.~2067--2080, 2022.

\bibitem{chen2021low}
Y.~Chen, X.~Xiao, C.~Peng, G.~Lu, and Y.~Zhou, ``Low-rank tensor graph learning
  for multi-view subspace clustering,'' {\em IEEE Transactions on Circuits and
  Systems for Video Technology}, vol.~32, no.~1, pp.~92--104, 2022.

\bibitem{xie2016Embedding}
J.~Xie, R.~B. Girshick, and A.~Farhadi, ``Unsupervised deep embedding for
  clustering analysis,'' in {\em Proceedings of the International Conference on
  Machine Learning}, vol.~48, pp.~478--487, 2016.

\bibitem{guo2017improved}
X.~Guo, L.~Gao, X.~Liu, and J.~Yin, ``Improved deep embedded clustering with
  local structure preservation.,'' in {\em Proceedings of the International
  Joint Conferences on Artificial Intelligence Organization}, pp.~1753--1759,
  2017.

\bibitem{sun2019self}
X.~Sun, M.~Cheng, C.~Min, and L.~Jing, ``Self-supervised deep multi-view
  subspace clustering,'' in {\em Asian Conference on Machine Learning},
  pp.~1001--1016, 2019.

\bibitem{zhu2019networks}
P.~Zhu, B.~Hui, C.~Zhang, D.~Du, L.~Wen, and Q.~Hu, ``Multi-view deep subspace
  clustering networks,'' {\em ArXiv: 1908.01978}, 2019.

\bibitem{wang2021Deep}
Q.~Wang, J.~Cheng, Q.~Gao, G.~Zhao, and L.~Jiao, ``Deep multi-view subspace
  clustering with unified and discriminative learning,'' {\em IEEE Transactions
  on Multimedia}, vol.~23, pp.~3483--3493, 2021.

\bibitem{cui2021Guided}
B.~Cui, H.~Yu, L.~Zong, and Z.~Cheng, ``Self-guided deep multi-view subspace
  clustering network,'' in {\em Proceedings of the {IEEE} International
  Conference on Multimedia and Expo}, pp.~1--6, 2021.

\bibitem{grill2020bootstrap}
J.-B. Grill, F.~Strub, F.~Altch{\'e}, C.~Tallec, P.~Richemond, E.~Buchatskaya,
  C.~Doersch, B.~Avila~Pires, Z.~Guo, M.~Gheshlaghi~Azar, {\em et~al.},
  ``Bootstrap your own latent: a new approach to self-supervised learning,''
  {\em Proceedings of the Advances in Neural Information Processing Systems},
  vol.~33, pp.~21271--21284, 2020.

\bibitem{hassani2020contrastive}
K.~Hassani and A.~H. Khasahmadi, ``Contrastive multi-view representation
  learning on graphs,'' in {\em Proceedings of the International Conference on
  Machine Learning}, pp.~4116--4126, 2020.

\bibitem{xu2022Multilevel}
J.~Xu, H.~Tang, Y.~Ren, X.~Zhu, and L.~He, ``Multi-level feature learning for
  contrastive multi-view clustering,'' in {\em Proceedings of the IEEE
  Conference on Computer Vision and Pattern Recognition}, 2022.

\bibitem{trosten2021reconsidering}
D.~J. Trosten, S.~Lokse, R.~Jenssen, and M.~Kampffmeyer, ``Reconsidering
  representation alignment for multi-view clustering,'' in {\em Proceedings of
  the IEEE/CVF Conference on Computer Vision and Pattern Recognition},
  pp.~1255--1265, 2021.

\bibitem{luo2018consistent}
S.~Luo, C.~Zhang, W.~Zhang, and X.~Cao, ``Consistent and specific multi-view
  subspace clustering,'' in {\em Proceedings of the International Joint
  Conferences on Artificial Intelligence Organization}, pp.~3730--3737, 2018.

\bibitem{liu2021Incomplete}
J.~Liu, X.~Liu, Y.~Zhang, P.~Zhang, W.~Tu, S.~Wang, S.~Zhou, W.~Liang, S.~Wang,
  and Y.~Yang, ``Self-representation subspace clustering for incomplete
  multi-view data,'' in {\em Proceedings of the {ACM} Multimedia Conference},
  pp.~2726--2734, 2021.

\bibitem{xie2020hyper}
Y.~Xie, W.~Zhang, Y.~Qu, L.~Dai, and D.~Tao, ``Hyper-laplacian regularized
  multilinear multiview self-representations for clustering and semisupervised
  learning,'' {\em {IEEE} Transactions on Cybernetics}, vol.~50, no.~2,
  pp.~572--586, 2020.

\bibitem{li2022preserved}
Z.~Li, C.~Tang, X.~Zheng, X.~Liu, W.~Zhang, and E.~Zhu, ``High-order
  correlation preserved incomplete multi-view subspace clustering,'' {\em
  {IEEE} Transactions on Image Processing}, vol.~31, pp.~2067--2080, 2022.

\bibitem{Kang2020Large}
Z.~Kang, W.~Zhou, Z.~Zhao, J.~Shao, M.~Han, and Z.~Xu, ``Large-scale multi-view
  subspace clustering in linear time,'' in {\em Proceedings of the
  International Joint Conferences on Artificial Intelligence Organization},
  pp.~4412--4419, 2020.

\bibitem{wang2022Guidance}
S.~Wang, X.~Liu, X.~Zhu, P.~Zhang, Y.~Zhang, F.~Gao, and E.~Zhu, ``Fast
  parameter-free multi-view subspace clustering with consensus anchor
  guidance,'' {\em {IEEE} Transactions on Image Processing}, vol.~31,
  pp.~556--568, 2022.

\bibitem{he2020momentum}
K.~He, H.~Fan, Y.~Wu, S.~Xie, and R.~Girshick, ``Momentum contrast for
  unsupervised visual representation learning,'' in {\em Proceedings of the
  IEEE/CVF Conference on Computer Vision and Pattern Recognition},
  pp.~9729--9738, 2020.

\bibitem{chen2020simple}
T.~Chen, S.~Kornblith, M.~Norouzi, and G.~Hinton, ``A simple framework for
  contrastive learning of visual representations,'' in {\em Proceedings of the
  International Conference on Machine Learning}, pp.~1597--1607, 2020.

\bibitem{niu2021Semantic}
C.~Niu and G.~Wang, ``{SPICE:} semantic pseudo-labeling for image clustering,''
  {\em ArXiv: 2103.09382}, 2021.

\bibitem{li2021Contrastive}
Y.~Li, P.~Hu, J.~Z. Liu, D.~Peng, J.~T. Zhou, and X.~Peng, ``Contrastive
  clustering,'' in {\em Proceedings of the {AAAI} Conference on Artificial
  Intelligence}, pp.~8547--8555, 2021.

\bibitem{zhong2021Graph}
H.~Zhong, J.~Wu, C.~Chen, J.~Huang, M.~Deng, L.~Nie, Z.~Lin, and X.~Hua,
  ``Graph contrastive clustering,'' in {\em Proceedings of the International
  Conference on Computer Vision}, pp.~9204--9213, 2021.

\bibitem{pan2021Graph}
E.~Pan and Z.~Kang, ``Multi-view contrastive graph clustering,'' in {\em
  Proceedings of the Advances in Neural Information Processing Systems},
  pp.~2148--2159, 2021.

\bibitem{xu2013survey}
C.~Xu, D.~Tao, and C.~Xu, ``A survey on multi-view learning,'' {\em arXiv
  preprint arXiv:1304.5634}, 2013.

\bibitem{yan2006graph}
S.~Yan, D.~Xu, B.~Zhang, H.-J. Zhang, Q.~Yang, and S.~Lin, ``Graph embedding
  and extensions: A general framework for dimensionality reduction,'' {\em IEEE
  Transactions on Pattern Analysis and Machine Intelligence}, vol.~29, no.~1,
  pp.~40--51, 2006.

\bibitem{nie2014projected}
F.~Nie, X.~Wang, and H.~Huang, ``Clustering and projected clustering with
  adaptive neighbors,'' in {\em Proceedings of the International Conference on
  Knowledge Discovery and Data Mining}, pp.~977--986, 2014.

\bibitem{asuncion2007uci}
A.~Asuncion and D.~Newman, ``Uci machine learning repository,'' 2007.

\bibitem{Zhan2019consensus}
K.~Zhan, F.~Nie, J.~Wang, and L.~Yang, ``Multiview consensus graph
  clustering,'' {\em IEEE Transactions on Image Processing}, vol.~28, no.~3,
  pp.~1261--1270, 2019.

\bibitem{Yang2019Split}
Z.~{Yang}, Q.~{Xu}, W.~{Zhang}, X.~{Cao}, and Q.~{Huang}, ``Split
  multiplicative multi-view subspace clustering,'' {\em IEEE Transactions on
  Image Processing}, vol.~28, no.~10, pp.~5147--5160, 2019.

\bibitem{liu2022Efficient}
S.~Liu, S.~Wang, P.~Zhang, X.~Liu, K.~Xu, C.~Zhang, and F.~Gao, ``Efficient
  one-pass multi-view subspace clustering with consensus anchors,'' in {\em
  Proceedings of the {AAAI} Conference on Artificial Intelligence}, 2022.

\bibitem{liu2021Cotraining}
J.~Liu, X.~Liu, Y.~Yang, X.~Guo, M.~Kloft, and L.~He, ``Multiview subspace
  clustering via co-training robust data representation,'' {\em IEEE
  Transactions on Neural Networks and Learning Systems}, 2021.
\newblock doi={10.1109/TNNLS.2021.3069424}.

\end{thebibliography}

\end{document}